\documentclass[10pt,twocolumn,letterpaper]{article}

\usepackage[pagenumbers]{cvpr} % To force page numbers, e.g. for an arXiv version

\usepackage{xcolor}
\usepackage{soul}
\usepackage{array}
\usepackage{multirow}
\usepackage{pdflscape}
\usepackage{afterpage}
\usepackage{svg}
\usepackage[accsupp]{axessibility} % Improves PDF readability for those with visual impairments.

\newcolumntype{C}[1]{>{\centering\arraybackslash}m{#1}}
\newcolumntype{L}[1]{>{\raggedright\arraybackslash}m{#1}}
\newcolumntype{M}[1]{>{\centering\arraybackslash}m{#1}}
\newcolumntype{R}[1]{>{\raggedleft\arraybackslash}m{#1}}

\usepackage{glossaries}
\usepackage{paralist}
\usepackage{algorithm}% http://ctan.org/pkg/algorithms
\usepackage{algpseudocode}% http://ctan.org/pkg/algorithmicx

\definecolor{cvprblue}{rgb}{0.21,0.49,0.74}
\usepackage[pagebackref,breaklinks,colorlinks,citecolor=cvprblue]{hyperref}

\def\paperID{7171} % *** Enter the Paper ID here
\def\confName{CVPR}
\def\confYear{2024}

\title{Explaining CLIP's performance disparities on data from blind/low vision users}

\author{Daniela Massiceti\textsuperscript{\dag}
\and
Camilla Longden\textsuperscript{\dag}
\and
Agnieszka S\l{}owik\textsuperscript{\dag}
\and
Samuel Wills\textsuperscript{$\diamond$}
\and
Martin Grayson\textsuperscript{\dag}\\\\
\textsuperscript{\dag}Microsoft Research
\and
Cecily Morrison\textsuperscript{\dag}\\\\
\textsuperscript{$\diamond$}The World Bank}

\begin{document}
\crefname{appendix}{App.}{appendices}
\maketitle
\newacronym{blv}{BLV}{blind and low vision}
\newacronym{vl}{VL}{vision-language}
\newacronym{lmm}{LMM}{large multi-modal model}
\newacronym{ols}{OLS}{Ordinary Least Squares}
\begin{abstract}
\vspace{-1em}
Large multi-modal models (LMMs) hold the potential to usher in a new era of automated visual assistance for people who are blind or low vision (BLV). Yet, these models have not been systematically evaluated on data captured by BLV users. We address this by empirically assessing CLIP, a widely-used LMM likely to underpin many assistive technologies. Testing 25 CLIP variants in a zero-shot classification task, we find that their accuracy is 15 percentage points lower on average for images captured by BLV users than web-crawled images. This disparity stems from CLIP's sensitivities to 1) image content (e.g.\! not recognizing disability objects as well as other objects); 2) image quality (e.g.\! not being robust to lighting variation); and 3) text content (e.g.\! not recognizing objects described by tactile adjectives as well as visual ones). We delve deeper with a textual analysis of three common pre-training datasets: LAION-400M, LAION-2B and DataComp-1B, showing that disability content is rarely mentioned. We then provide three examples that illustrate how the performance disparities extend to three downstream models underpinned by CLIP: OWL-ViT, CLIPSeg and DALL-E2. We find that few-shot learning with as few as 5 images can mitigate CLIP's quality-of-service disparities for BLV users in some scenarios, which we discuss alongside a set of other possible mitigations.
\vspace{-0.5em}
\end{abstract}    
\vspace{-1.5em}
\section{Introduction}
\label{sec:intro}

\begin{figure}[ht]
  \centering
  \scalebox{0.85}{
   \includegraphics[width=1\linewidth]{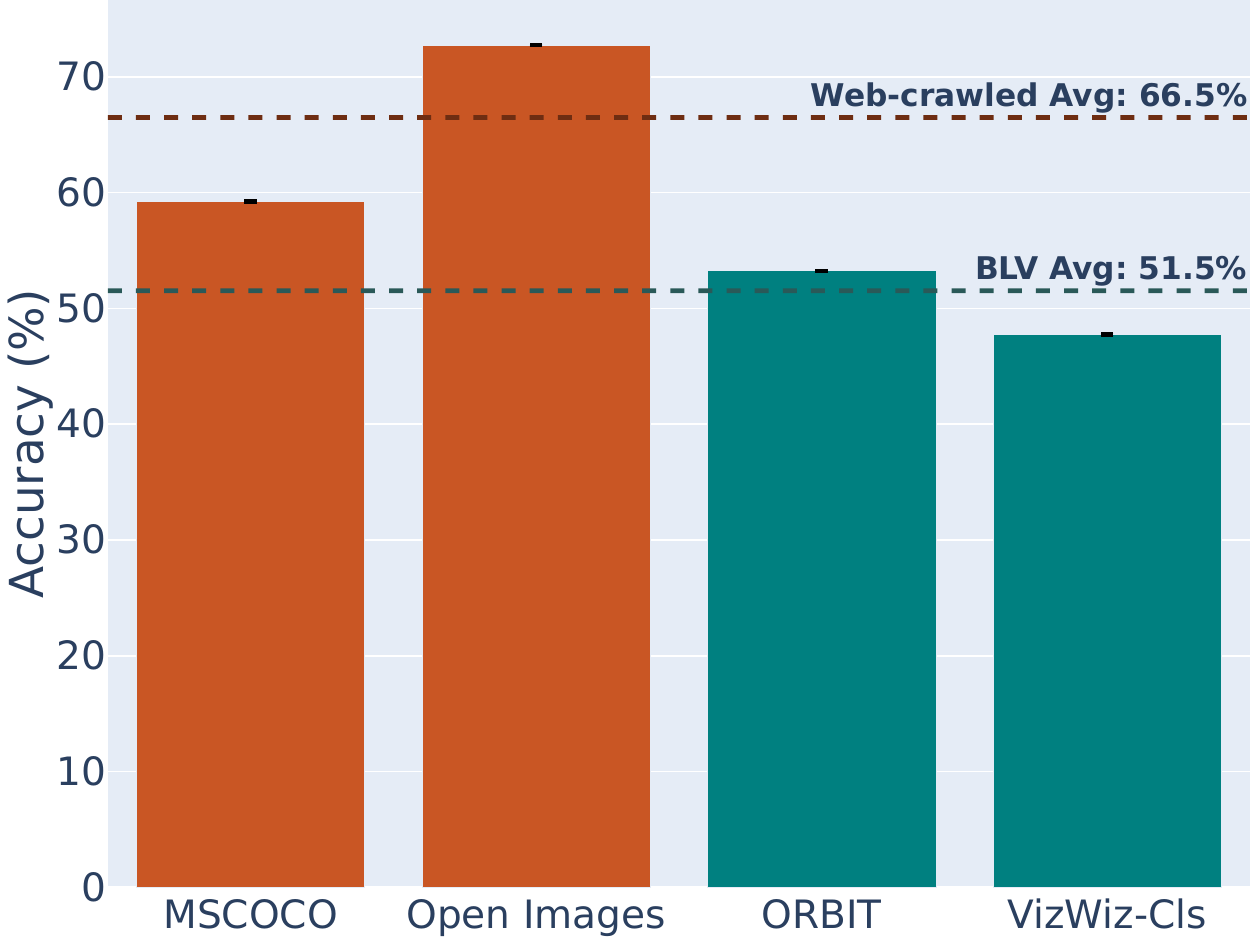}}
   \vspace{-0.5em}
   \caption{\textbf{CLIP's zero-shot object recognition accuracy is 15 percentage points lower in images from \acrshort{blv} users (ORBIT, VizWiz-Classification) versus web-crawled images (MSCOCO, Open Images).} Average accuracy (with 95\% c.i.) in a standardized zero-shot image classification task is reported over 80-100K images per dataset for 25 CLIP variants. %(\ie architecture size, pre-training dataset).
   }
   \label{fig:blv-vs-webcrawled}
   \vspace{-1.5em}
\end{figure}
AI-based applications hold the potential to help people who are \gls{blv} with everyday visual tasks~\cite{seeingai,googlelookout}.
However, the popularity of video-calling services like Be My Eyes~\cite{bemyeyes} suggest that human assistance is still often required due to the wide set of assistance tasks~\cite{morrison2023understanding} and varying quality of \gls{blv} images~\cite{chiu2020assessing,bafghi2023new}.
Recent advances in \glspl{lmm}~\cite{gptvmodelcard,radford2021learning,dai2023instructblip} could potentially address these challenges, enabling a new era of automated visual assistance as highlighted by the early partnership between Open AI and Be My Eyes~\cite{bemyeyesopenai}.

Despite the opportunity, little work has evaluated how well \glspl{lmm} perform on data from \gls{blv} users.
Performance disparities have been identified for other user groups~\cite{agarwal2021evaluating,wang2022measuring,luccioni2023stable,naik2023social,radford2021learning,rombach2022high} but the evidence for \gls{blv} users is either anecdotal~\cite{gptvmodelcard} or not specific to \emph{large} multi-modal models~\cite{bafghi2023new}. 
%that have been pre-trained on massive-scale data.
% %
Since \gls{blv} users are likely to be one of the biggest beneficiaries of \glspl{lmm}, often in productivity- and safety-critical situations, it is important to extend studies to this group.

To address this, we systematically evaluate CLIP, a widely used \gls{lmm} with 8700+ citations and 24M+ downloads\footnote{Statistics taken from Google Scholar and \href{https://huggingface.co/openai}{OpenAI's Hugging Face Hub} (for CLIP ViT-L/14, ViT-B/32 and ViT-B/16) on 23 October 2023.}, on data from \gls{blv} users.
CLIP's rich embeddings and strong zero-shot capabilities have led to it underpinning a wide range of downstream tasks including image classification~\cite{radford2021learning}, object detection~\cite{minderer2022simple,minderer2023scaling}, semantic segmentation~\cite{lueddecke22cvpr}, image captioning~\cite{tewel2022zerocap,su2022language} and video recognition~\cite{lin2022frozen}.
It has also been used to create large-scale datasets~\cite{schuhmann2021laion400m,schuhmann2022laion5B,gadre2023datacomp,lin2021clear} and evaluation metrics~\cite{hessel2021clipscore,park2021benchmark}.
As CLIP's pre-trained parameters are often used directly, poor performance can have wide-ranging implications for downstream assistive applications that use them.

We investigate CLIP's performance on \gls{blv} data along three dimensions: image content, image quality, and textual content.
Visual content considers how well CLIP can recognize \gls{blv}-specific objects, such as guide canes.
Visual quality assesses robustness to quality variations that characterize \gls{blv} images, such as blur and atypical framing~\cite{chiu2020assessing}.
Textual content examines performance on tactile descriptive words used by \gls{blv} users in contrast to visual ones, for example ``plastic'' versus ``yellow''.
We study each dimension in the context of a zero-shot image classification task, providing a worst-case estimate on how well CLIP will serve downstream assistive applications if used out-of-the box.

Overall, we find that CLIP's zero-shot classification accuracy is 15 percentage points lower on \gls{blv} images compared to web-crawled images across 25 CLIP variants.
These variants span architecture size (ViT-B/16 to ViT-g/14), pre-training dataset (WIT~\cite{radford2021learning},\! LAION~\cite{schuhmann2021laion400m,schuhmann2022laion5B},\! DataComp/CommonPool~\cite{gadre2023datacomp}) and pre-training dataset size (80M to 3.8B).
On deeper inspection, underperformance stems from CLIP:
\begin{inparaenum}[1)]
    \item recognizing disability objects less well than non-disability ones, with 25 percentage points lower accuracy;
    \item being sensitive to image quality, particularly occlusion and lighting issues; and 
    \item recognizing objects described by material less well than color, with discrepancies of 7 percentage points. 
\end{inparaenum}
In all cases, a larger pre-training dataset or architecture does not lead to parity.

%To further understand our results, we present an analysis of two large web-crawled datasets used to pre-train CLIP as well as an example-based analysis.
To further understand our results, we examine the upstream source and downstream impact of these disparities.
First, we conduct a textual analysis of the captions in LAION-400M/2B and DataComp1B and find that disability objects and materials are mentioned $\sim$17x and $\sim$4x less frequently than non-disability objects and colors, respectively.
Second, we find performance disparities on \gls{blv} data persist in three downstream models that use CLIP: OWL-ViT~\cite{minderer2022simple} for object detection, CLIPSeg~\cite{lueddecke22cvpr} for semantic segmentation, and DALL-E2~\cite{ramesh2022hierarchical} for text-to-image generation.
%
%We find that a few-shot approach with as few as 5 images can reduce these disparities in some scenarios.
%
We close by discussing a set of possible mitigations, including few-shot model adaption and application-level solutions, toward making automated visual assistance for \gls{blv} users more equitable.

In summary, our work contributes to the literature on how \glspl{lmm} perform for users in the margins, specifically highlighting how CLIP may underperform for \gls{blv} users if integrated into assistive applications.
Our contributions are:
\begin{compactitem}
\setdefaultleftmargin{0.5em}{}{}{}{}{}
    \item An empirical study of CLIP's performance on \gls{blv} image content, image quality and textual content.
    \item The first quantification of \gls{blv} content representation in LAION-400M, LAION-2B, and DataComp-1B.
    \item An example-based analysis that illustrates how performance disparities on \gls{blv} data persist in three downstream models that use CLIP.
    %\item A set of mitigations to address the performance disparities of large \gls{vl} models on \gls{blv} data.
\end{compactitem}
\section{Related Works}

\paragraph{Large multi-modal models.}
\Glspl{lmm} now have impressive capabilities in analyzing and synthesizing images~\cite{radford2021learning,yuan2021florence,jia2021scaling,gptvmodelcard,dai2023instructblip,alayrac2022flamingo,chen2022pali}.
%have the potential to revolutionize many assistive \gls{blv} technologies.
%
Contrastive models~\cite{radford2021learning,yuan2021florence,jia2021scaling}, a prominent sub-class, learn joint image and text embeddings by training on massive web-crawled data using a contrastive loss~\cite{chen2020simple, oord2018representation}.
They are unique in their architecture scale, and in the way they are trained on web-crawled data in an unsupervised manner.
Unlike previous models, the rich embeddings they learn are leveraged by a wide range of downstream models -- either directly~\cite{radford2021learning,fang2021clip2video}, or as part of a larger system~\cite{minderer2022simple,minderer2023scaling,lueddecke22cvpr,wang2022cris,tewel2022zerocap,su2022language,barraco2022unreasonable,mokady2021clipcap,lin2022frozen,tang2021clip4caption,ramesh2022hierarchical,nichol2021glide}.
%
%It is, therefore, critical to understand how well these embeddings represent image and text data captured by \gls{blv} users, given the likelihood that they will be integrated in downstream assistive applications in the future.

\vspace{-1em}
\paragraph{\glspl{lmm} and fairness.}
\Glspl{lmm} are known to have social biases across gender, race, age, and geographic location~\cite{agarwal2021evaluating, wang2022measuring,luccioni2023stable,naik2023social}.
CLIP, for example, has been shown to classify people of color as non-human and males as criminal more often than white people and females, respectively~\cite{agarwal2021evaluating}.
Some works have studied these representational harm for people with disabilities, however only in natural language~\cite{hutchinson2020social}.
%,are a function of the skew in the pre-training data~\cite{bird2020fairlearn,gebru2021datasheets}.
%
Quality-of-service harms arise when an application underperforms or fails for a particular user group~\cite{bird2020fairlearn,corbett2023measure,wan2021modeling} -- \eg a facial recognition system that does not detect women with darker skin tones~\cite{buolamwini2018gender}.
These can be systematically identified and mitigated through disaggregated reporting of a model's performance~\cite{barocas2021designing,nushi2018towards}.
This has not been well studied for people with disabilities generally or \gls{blv} people specifically, with the evidence either anecdotal (\eg GPT-4Vision model card~\cite{gptvmodelcard}) or not specific to \glspl{lmm}~\cite{bafghi2023new}.

\section{Methodology}
\label{sec:prelims}

Our work investigates CLIP's robustness to image and text data from \gls{blv} users in the context of a zero-shot image classification task.
This provides a worse-case estimate of how CLIP will perform out-of-the-box in downstream assistive applications.
Here we describe the experimental set-up, CLIP variants, and datasets used in our analyses.

\subsection{Episodic zero-shot image classification}
\label{sec:prelims:zero-shot}

An image classifier selects which object $c\! \in\! \mathcal{C}$ is present in an image, where $\mathcal{C}$ is the set of possible object classes and $|\mathcal{C}|$ is the task's ``way".
A zero-shot classifier does this without seeing any training images of the classes beforehand.

Our first analysis compares CLIP's performance on different datasets (rather than the more typical multiple models on a single dataset), requiring our classification task set-up to be standardized across datasets.
We take inspiration from the episodic sampling used in meta-learning~\cite{finn2017model}: for each dataset $j$ annotated with $\mathcal{C}_j$ object classes, we sample $T$ fixed $N$-way classification tasks, where for each task we randomly sample $N$ classes from $\mathcal{C}_j$.
For each task, we randomly sample $M$ test images per class.
% (\ie images labeled with that class as the ground truth).
%
The classification accuracy is then computed for all $T$*$M$*$N$ images and the average (and 95\% confidence interval) is reported.
We repeat this for each dataset, with $T$, $N$ and $M$ held constant.
We use variations of this to compare CLIP's performance between object types (\cref{sec:image-content-robustness}) and text prompts (\cref{sec:textual-content-robustness}) with details provided in each section, respectively.

\subsection{Logistic Regression}
\label{sec:prelims:regression}

We also aim to understand which characteristics \emph{within} images and text affect CLIP's performance.
We use logistic regression, a common tool for hypothesis testing, to estimate the marginal effect of each characteristic on the model's accuracy.
This approach avoids the need for careful experimental set-up which controls for all factors except the variable of interest.
Logistic regression extends \gls{ols} regression to the case when the output variable is binary, as is our case where the model correctly identifies the ground-truth object or not.
Formally, we use:
\vspace{-0.5em}
\begin{align}
\label{eq:logit-regression}
\begin{split}
p(z_i) &= \frac{1}{1+e^{-z_i}} \\
%     &where \\
%z_i  &= \alpha_1 + \beta_1 X_i + \alpha_2 D_i + \beta_2 D_i X_i + \alpha_3 FE_i + \epsilon_i
\end{split}
\end{align}
where $z_i  = \alpha_1 + \beta_1 X_i + \alpha_2 D_i + \beta_2 D_i X_i + \epsilon_i$.
The output variable is $p(z_i) \in [0,1]$, the probability that the model correctly identifies the ground-truth object in image $i$, with 1 for correct, and 0 otherwise.
The explanatory variables are $X_i$, a vector of binary variables that encode whether a particular characteristic is present in image $i$, and $D_i$ is a binary variable indicating whether the ground-truth object is a disability object (\eg a guide cane).
The interaction term $\beta_2 D_i X_i$ measures whether the marginal effect of each characteristic in $X_i$ is compounded or mitigated for disability objects relative to non-disability objects. $\epsilon_i$ are residuals which are assumed homoskedastic and uncorrelated.

The coefficients $\alpha_1, \beta_1, \alpha_2, \beta_2$ are estimated through maximum likelihood. In \gls{ols} the coefficients directly represent the marginal effect of each $X_i$ variable on the dependent variable. In contrast, here they represent the marginal effect on the log-odds ratio, which is linear in $X_i$:
\vspace{-0.3em}
\begin{equation}
\label{eq:log-odds}
\ln{\left(\frac{p(z_i)}{1-p(z_i)}\right)} = \alpha_1 + \beta_1 X_i + \alpha_2 D_i + \beta_2 D_i X_i + \epsilon_i
\end{equation}
This makes the coefficients difficult to interpret so we instead report them as $\partial p / \partial x$, the marginal effect of each characteristic $x\! \in\! X$ on the model's probability of being correct, $p$.
We report the average of this marginal effect across all observations in the sample.
We interpret each effect through its sign, magnitude, and significance.
A negative sign means the model is less likely to be correct when that characteristic is present in an image -- on average and holding all other characteristics constant. 
Its magnitude measures the extent of this impact.
Its significance indicates its reliability based on a two-sided t-test that estimates the probability that the marginal effect is different from zero.

\subsection{CLIP variants}

We study 25 CLIP variants spanning architecture size, pre-training dataset, and pre-training dataset size (see~\cref{app:tab:clip-variants} for summary).
We focus on variants that use a Transformer~\cite{vaswani2017attention} and Vision Transformer (ViT)~\cite{dosovitskiy2020image} as the text and vision encoders respectively as they are  most widely used. 
Specifically, we consider ViT-B/16, ViT-B/32, ViT-L/14, ViT-H/14 and ViT-g/14 vision encoders with associated text encoders.
For datasets, we consider OpenAI's closed-source WIT~\cite{radford2021learning} and open-source LAION (80M/400M/2B)~\cite{jia2021scaling,schuhmann2021laion400m,schuhmann2022laion5B}, DataComp (S/M/L/XL)~\cite{gadre2023datacomp}, and CommonPool (S/M/L/XL)~\cite{gadre2023datacomp} with and without CLIP Score filtering~\cite{hessel2021clipscore}.
These span 80M-3.8B image-text pairs.
%All model checkpoints are taken from \href{https://github.com/mlfoundations/open_clip}{\texttt{open\_clip}}~\cite{ilharcogabriel2021} and summarized in~\cref{tab:app:clip-variants}.

We use CLIP as a zero-shot classifier by embeddding a task's class labels using its text encoder, and each task image with its vision encoder.
An image's prediction is taken to be the class whose embedding has the highest cosine similarity (after a softmax) with the image's embedding.

\subsection{Datasets}

Our analyses are based on two large-scale datasets captured by \gls{blv} users: ORBIT~\cite{massiceti2021orbit} and VizWiz-Classification~\cite{bafghi2023new}. 
Both datasets were collected through real-world assistive applications: a personalizable object recognizer app for ORBIT~\cite{morrison2023understanding}; and a visual question-answering app for VizWiz-Classification~\cite{bigham2010vizwiz}.
Both are therefore highly representative of typical \gls{blv} user data. 
We contrast these with two common web-crawled datasets -- MS-COCO~\cite{lin2014microsoft} and Open Images~\cite{kuznetsova2020open} -- which are typical of the data used to pre-train \glspl{lmm}, and widely used for benchmarking.
We consider only the test and validation sets of these datasets.
Below we provide descriptions of the \gls{blv} datasets, with the web-crawled datasets described in the appendix.

%\vspace{-1em}
\textbf{ORBIT}~\cite{massiceti2021orbit} contains 3,822 videos (2.68M frames) of 486 objects collected by 67 \gls{blv} users on their mobile phones.
%
%Each user selected 5-10 personal objects which they wanted a personalized recognizer to be able to locate for them.
%
For each object, users captured videos which show the object alone, and in a realistic scene alongside other items, which we call the Clean and Clutter datasets, respectively.
%These two video types lend themselves to two evaluation scenarios for a zero-shot image classifier: in the first, we can test it on clean frames, and in the second, on clutter frames. 
ORBIT Clean frames are annotated with 6 quality issues (\eg framing, blur) following the categories in~\cite{chiu2020assessing}.

\textbf{VizWiz-Classification}~\cite{bafghi2023new} contains 8,900 images from the original VizWiz dataset~\cite{gurari2018vizwiz}, a dataset of images taken by over 11,000 \gls{blv} users via a visual assistance mobile app~\cite{bigham2010vizwiz}. All images are annotated with 200 ImageNet object categories and the 6 quality issues of~\cite{chiu2020assessing} (including an extra ``other'' quality issue).

\section{Experimental Results}
\label{sec:experiments}

Our first finding is that CLIP's accuracy is 15.0 percentage points lower on \gls{blv} datasets (ORBIT and VizWiz-Classification) than web-crawled datasets (MS-COCO and Open Images) (see~\cref{fig:blv-vs-webcrawled}).
We use the standardized zero-shot set-up (see~\cref{sec:prelims:zero-shot}) and average the $T$*$N$*$M$ predictions per dataset from each of the 25 CLIP variants.
While the accuracy difference is less for larger CLIP architectures than smaller ones, no model achieves parity (see~\cref{app:fig:blv-vs-webcrawled-per-model}). 
In the best case, the gap is 6.7 percentage points (ViT-g/14, LAION-2B) while in the worst, it is 22.8 percentage points (ViT-B/32, DataComp-M).
This preliminary result hints at deeper issues. 
In the following sections, we aim to identify potential sources of this discrepancy and why it occurs.

\subsection{Robustness to image content from \gls{blv} users}
\label{sec:image-content-robustness}

\begin{figure}
    \centering
    \includesvg[width=\linewidth]{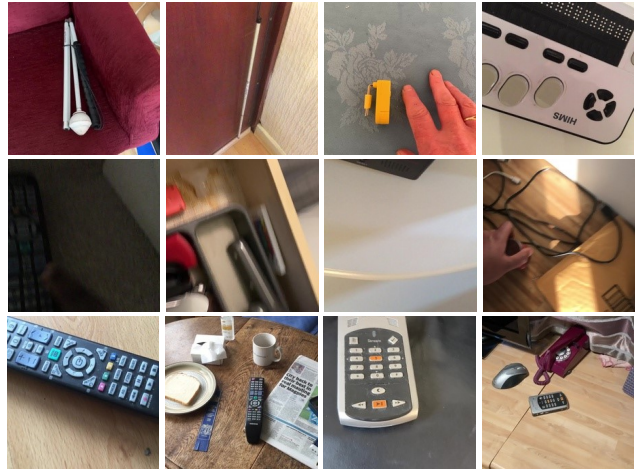}
    \vspace{-1.5em}
    \caption{\textbf{Examples from the ORBIT Dataset}. (top) Disability objects: guide canes, liquid level sensor, electronic Braille device. (middle) Quality issues typical in \gls{blv} images: underexposure, blur, camera viewpoint, and framing. (bottom) A remote control and a Victor Reader Stream in a clean and clutter frame.}
    \label{fig:orbit-examples}
    \vspace{-1.5em}
\end{figure}

To understand why accuracy is lower, we first examine \gls{blv} image content.
The \gls{blv} community uses a range of assistive objects, like guide canes and Braille displays~\cite{morrison2023understanding,massiceti2021orbit,kacorri2017people} (see~\cref{fig:orbit-examples}), which are not included in popular benchmarks~\cite{lin2014microsoft,imagenet15russakovsky,ridnik2021imagenet}.
We assess CLIP's performance on such ``disability'' objects versus more common objects.
%, after leveraging massive-scale web-crawled pre-training data. 
%We explore whether the diversity of massive-scale web-crawled data , through assessing CLIP's performance on disability versus non-disability objects.

We define disability objects as those that assist \gls{blv} people (\eg dog collar); exclusive disability objects as the subset exclusively used by \gls{blv} people (\eg guide cane); and non-disability objects as those used by everyone (\eg\!\! keys).
Three annotators categorized the ORBIT Clean and Clutter datasets\footnote{We do not consider VizWiz-Classification, as none of its 200 ImageNet labels are disability objects.} resulting in 55 disability, 42 exclusive disability, and 431 non-disability objects (see~\cref{app:sec:disability-objs} for lists).

\vspace{-1em}
\subsubsection{Disability objects are less well recognized than non-disability objects}
\label{sec:image-content:dis-vs-non-dis}
\begin{table}
\centering
\caption{\textbf{CLIP underperforms on disability and exclusive disability objects by significant margins compared to non-disability objects.} Zero-shot accuracy is averaged (with 95\% c.i.) over 27.5K images of each object type processed by each of the 25 CLIP variants. Experimental details in~\cref{sec:image-content:dis-vs-non-dis}.}
\vspace{-0.5em}
\scalebox{0.8}{
\label{tab:dis-vs-non-dis}
\begin{tabular}{l|c|c}
\toprule
\textbf{Object Category} & \textbf{ORBIT Clean} & \textbf{ORBIT Clutter} \\ 
\midrule
Excl. disability & $36.5\% \pm 0.1\%$ & $22.6\% \pm 0.1\%$ \\ 
Disability & $41.8\% \pm 0.1\%$ & $25.8\% \pm 0.1\%$ \\ 
Non-disability & \textbf{58.9\%} $\pm$ \textbf{0.1\%} & \textbf{50.9\%} $\pm$ \textbf{0.1\%} \\
% Excl. disability & $29.6\% \pm 0.1\%$ & $36.5\% \pm 0.1\%$ & $22.6\% \pm 0.1\%$ \\ 
% Disability & $33.8\% \pm 0.1\%$ & $41.8\% \pm 0.1\%$ & $25.8\% \pm 0.1\%$ \\ 
% Non-disability & $54.9\% \pm 0.1\%$ & $58.9\% \pm 0.1\%$ & $50.9\% \pm 0.1\%$ \\
\bottomrule
\end{tabular}}
\vspace{-1em}
\end{table}
We compare zero-shot classification accuracy between disability and non-disability objects using a variant of the episodic set-up described in~\cref{sec:prelims:zero-shot}.
Specifically, for each disability object we sample two $N$-way tasks with a ``target'' object and $N$-1 non-disability ``distractor'' objects.
The first task contains a disability target object and the second task contains a non-disability target.
The distractors are randomly sampled from the non-disability objects, each coming from a unique object cluster.
We repeat $T$ times for each disability object, sampling a pair of tasks with a different set of distractor objects and non-disability target object. 
For each task, we randomly sample $M$ frames of the target object, and ask CLIP to classify them from the task's $N$ possible objects. 
We report the average accuracy of all frames with a disability and a non-disability object as the target, respectively ($T$*$55$*$M$ each).
We also report the average accuracy over the subset of frames that are exclusive disability objects.
We use $T=5$, $N=20$, $M=100$.

Under this setting, we find that disability and exclusive disability objects have accuracies of 21.1 and 25.3 percentage points less than non-disability objects, respectively, on average across the ORBIT Clean dataset (see~\cref{tab:dis-vs-non-dis}). 
The gap widens by a further 3-4 percentage points when more realistic scenarios are presented from ORBIT Clutter. 
We find that the worst performing objects include Braille notetakers, talking book devices and liquid level indicators.

We also investigate the role of CLIP's pre-training dataset size on this finding.
%
%In~\cref{app:fig:orbit-clutter-dis-vs-non-dis-by-pretraining-size}, we plot the accuracy of non-disability, disability and exclusive disability objects on ORBIT Clutter with models ordered by their pre-training dataset size. 
%
We find that accuracy increases with pre-training dataset size generally, but the delta between non-disability and disability objects stays roughly constant (see~\cref{app:fig:orbit-clean-clutter-dis-vs-non-dis-by-pretraining-size}).
This suggests that web-crawling more data may not be enough to improve performance on potentially long-tailed objects.
%Indeed, we almost see a widening gap as the pre-training dataset size increases.
%
We see similar trends for increasing architecture sizes (see~\cref{app:fig:orbit-clean-clutter--dis-vs-non-dis-by-model-size}).

\vspace{-1em}
\subsubsection{Disability objects are under-represented in large-scale datasets compared to non-disability objects }
\label{sec:image-content:dataset-analysis-dis-vs-non-dis}

To better understand why more pre-training data does not improve performance on disability objects, we analyze the composition of three of CLIP's large-scale pre-training datasets for the presence of disability content -- LAION-400M~\cite{schuhmann2021laion400m}, LAION-2B~\cite{schuhmann2022laion5B}, and DataComp-XL~\cite{gadre2023datacomp} (also called DataComp-1B).
These datasets are used for pre-training \glspl{lmm} more broadly, with DataComp-XL achieving the highest accuracies on ORBIT.

Given the scale of the datasets, we conduct a text-based analysis of their captions as a more computationally tenable approach than analyzing their images.
We first extract all noun phrases that contain a physical object\footnote{A physical object traverses the entity $\rightarrow$ physical-entity $\rightarrow$ {object} $\rightarrow$ OR(artifact, whole, part, living-thing) hypernym path in WordNet~\cite{miller1995wordnet}.} from the captions, referred to as ``visual concepts''\footnote{We release these publicly at [REMOVED FOR REVIEW]}.
We then compute how prevalent ORBIT's disability and non-disability objects are contained in these visual concepts.
We use ORBIT to contextualize our previous results as it is a realistic representation of the types of objects important to \gls{blv} users, however, other object lists could be used.

To do this, we first group similar objects from the ORBIT dataset into higher-level clusters (\eg all guide canes). As each cluster could be described in several ways (\eg ``symbol canes'', ``guide canes''), we assign each two relevant synonyms.
This was expanded to 15 synonyms for disability objects based on initial experimentation, resulting in 222 disability object synonyms, and 312 non-disability synonyms overall.
We then count how many times each synonym appears within the visual concepts using string matching, allowing partial matches after simple pre-processing (see~\cref{app:sec:dataset-analysis} for details).

We find that disability objects occur 16-17x less frequently than non-disability objects across all three datasets (\cref{tab:dataset-analysis-dis-vs-non-dis}).
We compute this by normalizing the number of mentions by the number of synonyms for disability and non-disability objects, respectively, and taking their ratio.
%
%We also see that as the size of the LAION datasets increase, the frequency of disability objects mentions in fact decreases.
%
%Specifically, disability objects are mentioned in 0.0046\% of LAION-400M, but this proportion drops to 0.0025\% in LAION-2B.
%
%LAION-2B has 3.9x the number of disability mentions and 3.6x the number of non-disability mentions compared to LAION-400M. 
%
We also see that LAION-2B has 7x the number of noun phrases as LAION-400M, but \textless 4x the unique noun phrases, suggesting that it contains more of the same rather than new visual concepts (see~\cref{app:sec:dataset-analysis} for further statistics).
\begin{table}
    \centering
    \caption{\textbf{Disability objects occur 16-17x less frequently in the captions of popular large-scale image-text datasets compared to non-disability objects.} The mentions of 222 disability object synonyms and 312 non-disability synonyms were counted in noun phrases (NPs) extracted from these datasets. Details in~\cref{sec:image-content:dataset-analysis-dis-vs-non-dis}.}
    \vspace{-0.5em}
    \scalebox{0.75}{
    \begin{tabular}{L{2.5cm}|C{2.2cm}|C{2cm}|C{2.3cm}}
    \toprule
    & \textbf{LAION-400M} & \textbf{LAION-2B} & \textbf{DataComp-1B} \\
    \midrule
    Captions & 401,300,000 & 2,322,161,808 & 1,387,173,656 \\ \hline
    NPs & 384,468,921 & 2,737,763,447 & 1,342,369,058\\ \hline
    Unique NPs & 5,984,181 & 22,657,632 & 15,071,341 \\ \hline
    %Disability. obj synonyms & \makebox[0.2\linewidth]{222} & \makebox[0.2\linewidth]{222} & \\ \hline
    %Non-disability synonyms & \makebox[0.2\linewidth]{312} & \makebox[0.2\linewidth]{312} & \\ \hline
    Disability obj. mentions & 18,326 (0.0048\%) & 70,939 (0.0026\%) & 48,672 (0.0036\%)\\ \hline
    Non-disability obj. mentions & 425,046 (0.1106\%) & 1,550,043 (0.0566\%) & 1,126,356 (0.0839\%) \\ \hline
    %Raw non-dis/dis ratio & \makebox[0.2\linewidth]{23.6} & \makebox[0.2\linewidth]{22.0} \\ \hline
%    Disability obj. mentions per synonym & \makebox[0.2\linewidth]{80.15} & \makebox[0.2\linewidth]{311.56} & \\ \hline
%    Non-disability obj. mentions per synonym & \makebox[0.2\linewidth]{1344.63} & \makebox[0.2\linewidth]{4846.23} & \\ \hline
    \textbf{Normalized non-dis/dis ratio} & \textbf{16.8} & \textbf{15.6} & \textbf{16.5} \\
    \bottomrule
    \end{tabular}}
    \label{tab:dataset-analysis-dis-vs-non-dis}
    \vspace{-1em}
\end{table}

\vspace{-1em}
\subsubsection{A few-shot approach can \emph{sometimes} reduce the disability and non-disability accuracy gap}
\label{sec:image-content:few-shot}

As CLIP is also known to be a good few-shot learner~\cite{song2022clip}, we investigate whether providing several examples of an object can equalize performance between disability and non-disability content.
We integrate a ProtoNets approach~\cite{snell2017prototypical} with the ``distractor'' set-up described in~\cref{sec:image-content:dis-vs-non-dis}, using embeddings directly from CLIP's vision encoder\footnote{We note that this few-shot set-up does not use CLIP's text encoder.}.
Specifically for each disability object, we sample pairs of $N$-way tasks in the same way, except now we additionally sample $K$ training shots of each class which we use to compute the class prototypes.
As before, we evaluate the model on $M$ test images for the disability and non-disability target object in each task pair, with the prediction taken to be the closest prototype.
We consider $K=[5,10,20,40]$.
We compare this to recent methods~\cite{pratt2023does,menon2022visual} which improve CLIP's zero-shot performance by embedding LLM-generated descriptions of objects (rather than just the raw labels). We use GPT-4 as the LLM and the same generation hyperparameters as~\cite{menon2022visual,pratt2023does}.

%findings
We find that augmenting CLIP with LLM-generated object descriptions~\cite{pratt2023does,menon2022visual} outperforms vanilla CLIP (0-shot) which just embeds the raw object labels, but not a few-shot approach (5-shot) which embeds a few image examples of each object (see~\cref{tab:orbit-clean-clutter-few-shot-dis-vs-non-dis-delta}).
This holds for both the ORBIT Clean and Clutter datasets. 
Crucially, the accuracy gap between disability and non-disability objects is lowest with a few-shot approach, though this accuracy gap quickly saturates, with no significant gains coming from more than 5 shots (see~\cref{app:fig:orbit-all-few-shot-by-num-shots}).
We also note that while a few-shot approach can reduce the accuracy gap to 2\% in the simple images from ORBIT Clean, it is less effective in the more realistic images from ORBIT Clutter, with disability objects performing 14-15\% points worse than non-disability objects, even when scaled to 40 shots (see~\cref{app:fig:orbit-clutter-few-shot-by-num-shots}).
\begin{table}
    \centering
    \vspace{-0.5em}
    \caption{\!\!{\textbf{A few-shot method using ProtoNets~\cite{snell2017prototypical} (5-shot) achieves the highest accuracy and lowest accuracy gap between disability and non-disability objects, versus vanilla CLIP (0-shot) and CLIP with LLM-generated object descriptions~\cite{menon2022visual,pratt2023does}}.\! Averaged over 25 CLIP variants.}}
    \vspace{-0.8em}
    \scalebox{0.77}{
    \begin{tabular}  {p{2.1cm}|C{1cm}@{\hspace{0pt}}C{0.8cm}@{\hspace{1pt}}C{0.8cm}@{\hspace{1pt}}C{1cm}@{\hspace{1pt}}|C{1cm}@{\hspace{1pt}}C{1cm}@{\hspace{1pt}}C{0.6cm}@{\hspace{1pt}}C{1cm}}
        \hline
         \textbf{Obj type} & \multicolumn{4}{c}{\textbf{ORBIT Clean Acc (\%)}} & \multicolumn{4}{c}{\textbf{ORBIT Clutter Acc (\%)}}\\ & \textbf{0-shot} & \textbf{\cite{menon2022visual}} & \textbf{\cite{pratt2023does}} & \textbf{5-shot} & \textbf{0-shot} & \textbf{\cite{menon2022visual}} & \textbf{\cite{pratt2023does}} & \textbf{5-shot}  \\
         \hline
        Disability & 41.8 & 48.3 & 50.1 & \textbf{86.2} & 25.8 & 32.1 & 34.2 & \textbf{54.5} \\
        Non-disability & 58.9 & 57.0 & 57.0 & \textbf{88.3} & 50.9 & 50.2 & 49.4 & \textbf{69.1} \\
        \hline
        \textbf{Accuracy gap} & 17.1 & 8.7 & 6.9 & \textbf{2.1} & {25.1} & 18.1 & 15.2 & \textbf{14.6}\\ 
        \hline
    \end{tabular}}
    \label{tab:orbit-clean-clutter-few-shot-dis-vs-non-dis-delta}
    \vspace{-1.6em}
\end{table}
%\begin{figure}
%    \centering
%    \includegraphics[width=0.9\linewidth]{plots/orbit-clean-clutter-few-shot-dis-vs-non-dis-delta.pdf}
%    \caption{\textbf{A few-shot approach (ProtoNets~\cite{snell2017prototypical}) can effectively reduce the accuracy gap between disability and non-disability objects, but not for realistic, cluttered images.} Bars represent the delta between disability and non-disability object few-shot accuracy, averaged (with 95\% c.i.) over all test frames for each shot setting ($K\!\!=\!\![5,10,20,40])$. $K\!\!=\!\!0$ is equivalent to the zero-shot setting in~\cref{sec:image-content:dis-vs-non-dis}. Experimental details in~\cref{sec:image-content:few-shot}.}
%    \label{fig:orbit-clean-clutter-few-shot-dis-vs-non-dis-delta}
%    \vspace{-1em}
%\end{figure}

Furthermore, a few-shot approach is only effective as a mitigation if CLIP is pre-trained on a large enough dataset.
%
%\cref{app:fig:orbit_clean-few-shot-by-dataset-size} shows that the difference in CLIP's few-shot accuracy between disability and non-disability objects reduces as the pre-training dataset increases.
%
We find that for pre-training datasets of less than 100M examples, the accuracy difference is 3-4x larger than that for 100-1000M examples, and 9-10x larger than that for 1B+ examples (see~\cref{app:fig:orbit-clean-few-shot-by-dataset-size,app:fig:orbit-clutter-few-shot-by-dataset-size}).
These factors are roughly constant across the number of shots. 
Overall, this speaks to the power of large-scale pre-training, even if a small amount of extra effort is required.

%%%%%%%%%%%%%%%%%%%%%%%%%%%%%%%%%%%%%%%%%%%%%%%%%%%%%%%%%%%%%%%%%%%%%%%
\subsection{Robustness to image quality from \gls{blv} users}
\label{sec:image-quality-robustness}

Images captured by \gls{blv} users are of more variable quality than those captured by sighted users.
These issues include atypical framing, camera blur, camera viewpoint (rotation), occlusion, overexposure, and underexposure~\cite{chiu2020assessing,kacorri2017people}, which are annotated in the ORBIT Clean and VizWiz Classification datasets.
We run the standardized zero-shot set-up (see~\cref{sec:prelims:zero-shot}) on these datasets for all CLIP variants.
%To understand the impact of these quality issues on performance, we run the standardized zero-shot classification set-up (see~\cref{sec:prelims:zero-shot}) on the ORBIT Clean and VizWiz Classification datasets which have quality issue annotations for each image.
%
We then use the statistical tools described in~\cref{sec:prelims:regression} to disentangle the marginal effect of each quality issue on model performance, both in general and for disability objects specifically.
For ORBIT, we treat $X_i$ as a binary vector indicating the presence of five quality issues\footnote{We combine over- and underexposure into a joint ``lighting issue'' due to low incidence rates of each of these issues.} in image $i$, $D_i$ as a binary indicating the presence of an exclusive disability object, and $D_iX_i$ as the interactions between them.
For VizWiz, we encode seven quality issues (including the ``other'' category) in $X_i$, but exclude $D_i$ or $D_iX_i$ as VizWiz labels do not include disability objects.

% Our findings show that the accuracy of all models is reduced by quality issues, with overexposure, underexposure, blur and occlusion associated with the largest reduction in accuracy on average.
% %
% Accuracy is lower for disability objects than non-disability objects on average, cross-validating the findings seen in~\cref{sec:image-content:dis-vs-non-dis}.
% %
% However, disability objects are impacted no more than non-disability objects when they appear with quality issues.

\vspace{-1em}
\subsubsection{Blur, viewpoint, occlusion and lighting issues significantly reduce model accuracy.}

\begin{figure}
    \centering
    \includegraphics[width=\linewidth]{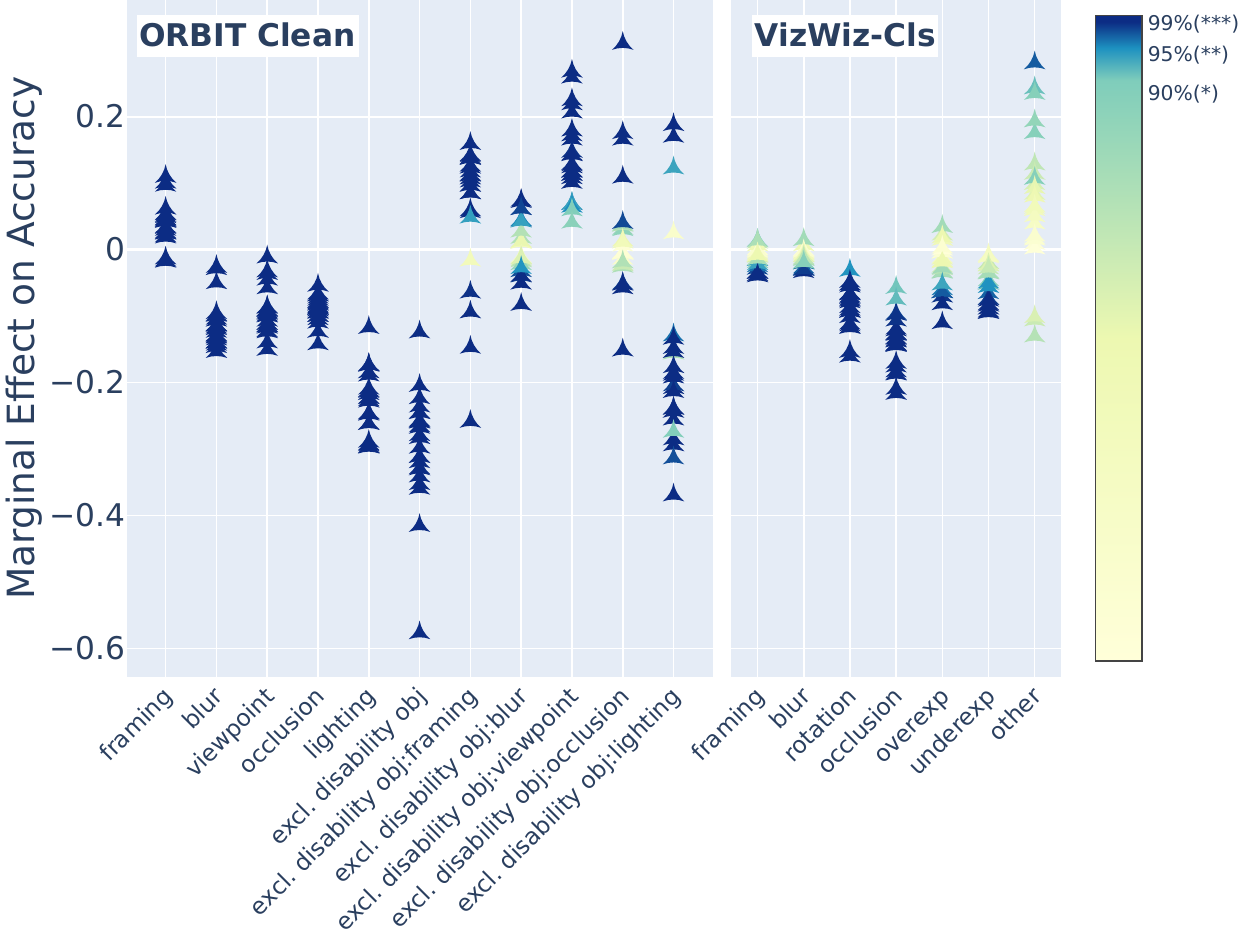}
    \vspace{-1.5em}
    \caption{\textbf{Blur, viewpoint/rotation, occlusion and lighting issues all have large negative marginal effects on model accuracy, with high statistical significance, but these are not compounded for exclusive disability objects.} Each dot represents a CLIP variant, with its color showing the significance level.}
    \label{fig:orbit-clean-vizwiz-classification-reg}
    \vspace{-1em}
\end{figure}

In~\cref{fig:orbit-clean-vizwiz-classification-reg}, we show that the marginal effects of blur, viewpoint (rotation), occlusion, and lighting issues on model accuracy are negative, large, and statistically significant for most models.
All else equal, blur reduces model accuracy by 11 percentage points and 1 percentage point in the ORBIT and VizWiz datasets, respectively, on average. 
Viewpoint issues by 9 and 8 percentage points on each dataset respectively; occlusion by 9 and 14 percentage points; and lighting issues by 23 and 8 percentage points.
We note that these effects are cumulative meaning that the impact on model accuracy is summed if multiple issues occur in the same image.
We also note that pre-training on larger datasets, in general, does not guarantee robustness (\eg variants pre-trained on LAION-2B, one of the largest datasets, are negatively affected by viewpoint and occlusion issues by 3-12 and 8-19 percentage points, respectively).
We include the raw marginal effects in~\cref{app:tab:orbit-vizwiz-reg-b16,app:tab:orbit-vizwiz-reg-b32,app:tab:orbit-vizwiz-reg-l14}.

Framing issues in the ORBIT dataset stand as the exception, with the marginal effect being positive and statistically significant.
This can be explained by how the ORBIT videos were collected. To orient the camera, \gls{blv} users were instructed to hold it close to the object initially, and then move away.
So, the initial frames in the video tend to be at close range -- an easier recognition task -- but also have framing issues.
This is supported by the VizWiz results where framing issues, which occur at further distances from the object, have a negative marginal effect on accuracy.

\vspace{-1em}
\subsubsection{The impact of quality issues is typically not worse for disability compared to non-disability objects.}

\cref{fig:orbit-clean-vizwiz-classification-reg} further shows that accuracy is 29 percentage points lower for exclusive disability objects than non-disability objects in the ORBIT Clean dataset, on average across all models, supporting the findings in~\cref{sec:image-content:dis-vs-non-dis}.
The marginal effect of a quality issue, however, typically affects disability objects no worse than non-disability ones.
This can be seen by comparing the net effect of a quality issue on each object type.
Let the baseline be the accuracy for non-disability objects.
The accuracy for a disability object with no quality issues will be 29 percentage points lower.
Introducing occlusion will reduce the accuracy for non-disability objects by 9 percentage points on average.
For disability objects, occlusion will reduce accuracy by this, plus the marginal effect of the interaction term (+2 percentage points), for a net effect of -7.
The positive and significant interaction term indicates that having an occlusion issue and being a disability object has an effect that is slightly less than the sum of its parts.
The only exception is overexposure issues, which do compound if they co-occur with a disability object.

\subsection{Robustness to language used by \gls{blv} users}
\label{sec:textual-content-robustness}

Assistive applications are likely to leverage the multi-modal capabilities of \glspl{lmm}, so it is important to understand how CLIP performs on the range of language used by \gls{blv} people.
For example, \gls{blv} users commonly use tactile rather than visual words to describe their objects \cite{morrison2023understanding}. 
In this section, we study one instantiation of this -- CLIP's robustness to recognizing objects described by their color, ``yellow mug'', versus their material, ``plastic cup''.

% methodology
To do this, three annotators manually labeled the ORBIT validation and test objects (208 objects) with a color and a material\footnote{We assigned up to 2 adjectives per object in some cases where objects were multiple colors or materials.}.
Each adjective was selected from a pre-defined list of 20 colors and 23 materials (see~\cref{app:sec:colors-materials}).
A text prompt was then created for each object using the template ``\emph{\textless adjective\textgreater{} \textless object\_name\textgreater}", where \emph{\textless adjective\textgreater} was the object's color or material, and \emph{\textless object\_name\textgreater} was the noun extracted from the raw object label.
We use these templates -- referred to as color and material prompts -- to examine CLIP's sensitivity to different object descriptions.

\vspace{-1em}
\subsubsection{CLIP classifies objects more accurately when they are described by color rather than material}
\label{sec:text-content:color-vs-material}

We compute CLIP scores~\cite{hessel2021clipscore} between an image and four different prompt embeddings, for 100 randomly sampled images of each object in ORBIT Clean.
%
%The CLIP score is a scaled cosine similarity between an image and text embedding and measures how correlated or aligned the two embeddings are -- a score of 0 indicates no alignment, while a maximum score of 100 indicates strong alignment.
%
%s been shown to be highly correlated with human judgement.
We consider the color and material prompts, a lower bound containing just the object name, and an upper bound adding both color and material adjectives.
%``\emph{\textless color\textgreater \textless material\textgreater \textless object\_name\textgreater}''.
%
We expect that the lower bound prompt, which provides the least detail about the object, should align less strongly with the object's image embedding than the upper bound prompt, which provides the most specific detail. 
In~\cref{tab:orbit-clean-avg-clip-score-per-prompt-type}, however, we see this is not the case.
\begin{table}
    \centering
    \caption{\textbf{Describing an object by its color (rather than material, or color and material) leads to text embeddings that are most aligned with that object's image embeddings.} CLIP scores~\cite{hessel2021clipscore} between image and prompt embeddings are averaged (with 95\% c.i.) for 100 images per object per prompt type on ORBIT Clean.} %Models: All 25 CLIP variants.}
    \vspace{-0.5em}
    \scalebox{0.8}{
    \begin{tabular}{l|C{1.2cm}|C{1.6cm}|C{1.6cm}|C{1.7cm}}
    \toprule
         \textbf{Prompt} & \textbf{Obj. name} & \textbf{Material + obj. name} & \textbf{Color + obj. name} & \textbf{Color + material + obj. name}  \\ \hline
         \textbf{CLIP Score} & 24.07 $\pm$ 0.02 & 23.88 $\pm$ 0.02 & \textbf{25.20} $\pm$ \textbf{0.02} & 24.76 $\pm$ 0.02 \\
    \bottomrule
    \end{tabular}}
    \label{tab:orbit-clean-avg-clip-score-per-prompt-type}
    \vspace{-1em}
\end{table}
Rather, color prompts have the highest CLIP scores and material prompts the lowest.
%, with an average of $25.20\pm 0.02$.
%, with an average of $23.88\pm 0.02$.
%
Interestingly, the upper bound has a lower average CLIP score
%($24.76\pm0.02$) 
than the color prompt, suggesting that adding the object's material is harming alignment.

To quantify the impact of this on accuracy, we run the standard zero-shot set-up (\cref{sec:prelims:zero-shot}), embedding these textual prompts instead of the raw object labels.
We see that across all variants, CLIP classifies objects 7.1 percentage points more accurately when they are described by their color rather than their material (see~\cref{app:fig:orbit-clean-orbit-clutter-avg-acc-per-prompt-type-by-model}).

\vspace{-1em}
\subsubsection{Materials are under-represented in large-scale datasets compared to colors}
\label{sec:text-content:dataset-analysis:material}
\begin{table}[b]
    \centering
    \vspace{-1em}
    \caption{\textbf{Materials occur $\sim$4x less frequently than colors in the captions of popular large-scale image-text datasets.} The mentions of 20 colors and 23 materials were counted in the noun phrases extracted in~\cref{tab:dataset-analysis-dis-vs-non-dis}.}
    \vspace{-0.5em}
    \scalebox{0.75}{
    \begin{tabular}{L{2.6cm}|C{2.2cm}|C{1.9cm}|C{2.3cm}}
    \toprule
    & \textbf{LAION-400M} & \textbf{LAION-2B} & \textbf{DataComp-1B} \\
    \midrule
    %Colors & \makebox[0.2\linewidth]{20} & \makebox[0.2\linewidth]{20} & \\ \hline
    %Materials & \makebox[0.2\linewidth]{23} & \makebox[0.2\linewidth]{23} & \\ \hline
    Color mentions & 475,060 (0.12\%) & 1,756,102 (0.06\%) & 1,165,871 (0.09\%) \\ \hline
    Material mentions & 131,876 (0.03\%) &  513,014 (0.02\%) & 354,598 (0.03\%) \\ \hline
    \textbf{Norm'ed color/ material ratio} & \textbf{\makebox[0.2\linewidth]{4.1}} & \textbf{\makebox[0.2\linewidth]{3.9}} & \textbf{\makebox[0.2\linewidth]{3.8}} \\
    \bottomrule
    \end{tabular}}
    \label{tab:dataset-analysis-material}
    \vspace{-1.5em}
\end{table}
We further examine this result by measuring how frequently colors versus materials appear in the captions of LAION-400M, LAION-2B and DataComp-1B.
We use the extracted noun phrases from~\cref{sec:image-content:dataset-analysis-dis-vs-non-dis}, and count the number of times the 20 material and 23 color annotations are mentioned.
In~\cref{tab:dataset-analysis-material}, we see that colors are mentioned $\sim$4x more frequently than materials across both datasets, once normalized.
This helps to explains some of the results in~\cref{sec:text-content:color-vs-material}.
Taken together, this suggests that models pre-trained on these datasets may perform worse for \gls{blv} users who describe their objects by their material, with the potential that this may extend to other tactile-based descriptions.

\section{Example-based impact analysis}
\label{sec:example-based-analyis}

\cref{sec:experiments} broadly shows that CLIP is sensitive to image and textual data provided by \gls{blv} users in a zero-shot classification task.
We investigate whether these performance disparities persist in three downstream models that use CLIP -- OWL-ViT~\cite{minderer2023scaling}, CLIPSeg~\cite{lueddecke22cvpr}, and DALL-E2~\cite{ramesh2022hierarchical}.
We run our analysis on 180 \gls{blv} images which are systematically selected for 20 objects -- the 5 top- and bottom-performing disability and non-disability objects from the ORBIT dataset (see~\cref{app:sec:example-analysis} for full protocol).
% with 9 images per object
%
For space reasons, we include CLIPSeg results in~\cref{app:sec:example-analysis:seg}.

\vspace{-0.4em}
\subsection{Object detection with OWL-ViT}

\begin{figure}
    \centering
    \includesvg[width=\linewidth,inkscapelatex=false]{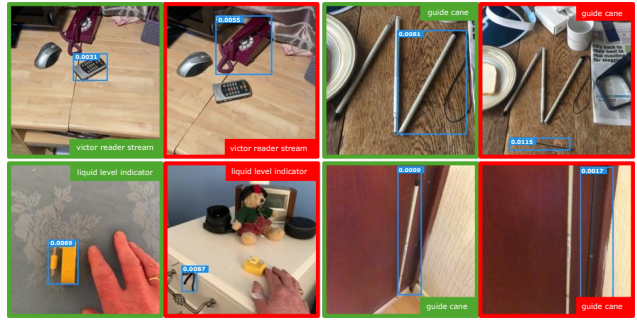}
    \caption{\textbf{OWL-ViT~\cite{minderer2022simple} detects disability objects less consistently than non-disability objects}. Disability objects are often mistaken for other objects, sometimes with higher confidence.}
    \label{fig:image-analysis:det1}
    \vspace{-1em}
\end{figure}

Object detection is already widely available in \gls{blv} assistive applications~\cite{seeingai,googlelookout}, and in future, many may rely on models that use CLIP, such as OWL-ViT~\cite{minderer2022simple}.
OWL-ViT predicts bounding boxes for objects specified in free-form text prompts. 
It does this by appending a bounding box regression and class-wise layer to CLIP's (pre-trained) encoders and then fine-tuning on an object detection dataset.
We run all 180 images through OWL-ViT (with a ViT-B/32 vision encoder) with the (cleaned) noun phrase extracted from the raw object label as the text prompt.
A team of three annotators then manually evaluated the detections.
We find:

\vspace{0em}
\textbf{Disability objects are less consistently detected than non-disability objects.}
Our results show that 6/10 non-disability objects were correctly detected (taken as the box with the highest confidence) in all 9 frames showing that object, compared to 3/10 disability objects.
In many of these failed frames, the model mistook the disability object for another object, often with a higher confidence (\cref{fig:image-analysis:det1}).
This behavior would have a large negative effect on the user experience of an object detection app.

\vspace{-0em}
\textbf{The model is less confident about disability object detections than non-disability object detections.}
In~\cref{tab:image-analysis:det2}, we see that OWL-ViT's confidence for the correct bounding box is $\sim$5x lower for disability objects compared to non-disability objects.
We see that incorrect boxes have similar confidence scores between disability and non-disability objects, which is expected.
See examples in~\cref{app:fig:image-analysis:det2}.
\begin{table}
    \centering
    \caption{\textbf{OWL-ViT~\cite{minderer2022simple}'s correct bounding box predictions have confidence scores that are $\sim$5x lower for disability than non-disability objects on average}. The confidence score of the predicted box per image is averaged (with 95\% c.i.) over 90 images for disability and non-disability objects, respectively.}
    \vspace{-0.5em}
    \scalebox{0.9}{
    \begin{tabular}{p{2.1cm}C{2.3cm}C{2.4cm}}
    \toprule
    \textbf{Object} & \textbf{Correct boxes} & \textbf{Incorrect boxes} \\
    \hline
    Dis. objs & 0.016 $\pm$ 0.008 & 0.008 $\pm$ 0.003 \\
    Non-dis. objs & 0.084 $\pm$ 0.030 & 0.008 $\pm$ 0.003 \\
    \bottomrule
    \end{tabular}}
    \label{tab:image-analysis:det2}
    \vspace{-2em}
\end{table}

\subsection{Text-to-image generation with DALL-E2}

\begin{figure}[b]
    \centering
    \vspace{-1em}
    \includegraphics[width=\linewidth]{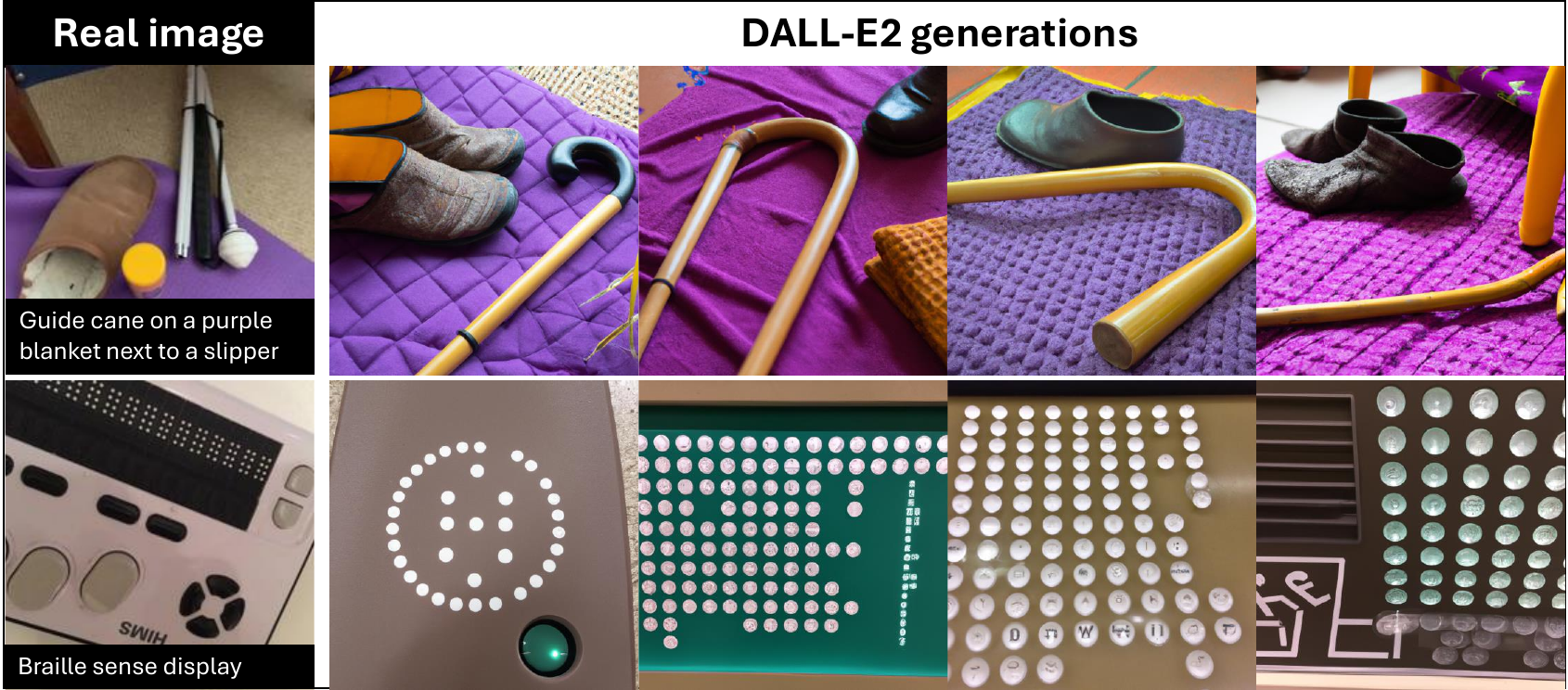}
    \vspace{-1.5em}
    \caption{\textbf{DALL-E2~\cite{ramesh2022hierarchical} either defaults to common objects or fabrications when prompted with disability objects like guide canes and electronic Braille devices}. Instead, it generates high-quality images of non-disability objects (see~\cref{app:fig:image-analysis:gen1a}).}
    \label{fig:image-analysis:gen1}
    %\vspace{-1em}
\end{figure}

DALL-E2~\cite{ramesh2022hierarchical} also uses CLIP: during training its decoder is conditioned on image embeddings from frozen CLIP. 
We investigate the downstream impacts of this by examining if DALL-E2 can generate disability content.
We create two prompts for each of the 20 objects using the templates:
\begin{inparaenum}[i)]
    \item ``\emph{\textless object\_name\textgreater}''
    \item ``\emph{\textless object\_name\textgreater{} on \textless surface\textgreater{} next to a \textless adjacent-object\textgreater}''.
\end{inparaenum}
The object name was the object label's cleaned noun phrase, and the surface/adjacent object was chosen to match a randomly sampled clutter image of that object (see~\cref{app:sec:example-analysis} for details). 
%
%In cases where other objects were not present, no adjacent objects were included in the caption.
%
Three annotators then manually evaluated four generations from DALL-E2 per prompt.
A generated image was considered correct if it contained the object specified in the prompt. We find:

\textbf{Generations of disability objects are more likely to be incorrect compared to non-disability objects.}
%For disability objects, 17/80 images were marked as "Yes".
%For non-disability objects, 66/80 images were marked as "Yes". 
%In the generations, not a single picture of a guide cane, braille reader, liquid level indicator or victor stream reader was generated successfully.
DALL-E2 correctly generated the object in the prompt for 18/80 images of disability objects, versus 74/80 images of non-disability objects.
For some disability objects, no generations contained a valid representation of the object -- including guide canes, electronic Braille devices, and liquid level indicators (see~\cref{fig:image-analysis:gen1,app:fig:image-analysis:gen1}).
In these cases, the generations either defaulted to a more common object (\eg a walking stick for ``guide cane'') or fabricated an object entirely (\eg random dot patterns for ``Braille sense display'', colorful thermometers for ``liquid level sensor'').
%The model did not generate any correct visualizations of Braille devices.
%
It also failed to generate specific instances of assistive devices (\eg ``Victor Reader Stream'', a talking book device, resulted in images of books or river streams).
In contrast, DALL-E2 generates highly realistic of non-disability objects (see~\cref{app:fig:image-analysis:gen1a}).

\vspace{-0.5em}
\section{Discussion}
\label{sec:discussion}

Our evaluation of CLIP reveals that it consistently underperforms on \gls{blv} data across visual content, visual quality, and textual content, irrespective of architecture size, pre-training dataset, or pre-training dataset size.
We discuss mitigation strategies to make \glspl{lmm} more equitable for \gls{blv} users and marginalized groups more generally.

Our results suggest that the performance disparities come in part from the distribution shift between web-crawled and \gls{blv} user data.
This highlights the importance of systematic reporting of the contents of large-scale datasets used for pre-training, in the spirit of datasheets for datasets~\cite{gebru2021datasheets}.
Our analysis in~\cref{sec:image-content:dataset-analysis-dis-vs-non-dis,sec:text-content:dataset-analysis:material} provides a starting point, but this should be extended to other datasets and marginalized content.
With the data composition known, mitigation strategies can then be developed.
For example, assistive device websites and disability dataset platforms like IncluSet~\cite{incluset} could explicitly be crawled.

We also show that a few-shot approach can mitigate performance disparities relating to image content -- a more cost-effective alternative than re-training a \gls{lmm}.
The few-shot model adaptation could be done when the \gls{lmm} is developed, when the application is developed, or by the end-users themselves as part of a teachable paradigm~\cite{massiceti2021orbit,kacorri2017people}.
Each of these options is an open research question with the need to more deeply explore interaction paradigms and light-weight model adaptation techniques~\cite{hu2021lora,basu2023strong}.

Finally, application-level mitigations should also be considered.
For \gls{blv} users, auxiliary models could support users to reduce image variance, helping them stabilize the camera or alerting about the lighting conditions, for example.
We could also leverage data augmentation techniques that are personalized to individual users or user groups.
For \gls{blv} users who tend to take blurry images, for example, we could automatically inject blur into the few-shot images so that the model becomes more robust to this quality issue.

The  findings in this paper prompt a critical look at the development cycle of current \glspl{lmm}.
Greater transparency and disaggregation in dataset reporting is needed, regardless of the proprietary nature of a dataset.
Future work should also explore lightweight model adaption techniques that allow application developers and users to bring equity to their experiences.
We must continue to work with marginalized communities -- ``nothing about us without us'' – to equalize the benefit of \glspl{lmm} and their extraordinary capabilities.

{
    \small
    \bibliographystyle{ieeenat_fullname}
    \bibliography{main}
}

\appendix
\clearpage
\newpage
%\setcounter{page}{1}
%\maketitlesupplementary
\counterwithin{figure}{section}
\counterwithin{table}{section}

\section{Extended experimental details}
\subsection{Datasets}

\paragraph{Open Images.}
The Open Images V7 dataset~\cite{kuznetsova2020open} contains 61.4M images with image-level labels spanning 20.6K object classes.
The images are web-crawled from Flickr and the classes include items of clothing, food types, animals, vehicles and more. 
We motivate the choice of this dataset because of its scale and diversity, and because it is widely used for training and benchmarking models within the computer vision community.
We only sample images from the validation and test splits that have been verified by humans to contain the labeled object (i.e. all false positives are removed).
This is 390,797 validation and 1,319,751 test images, respectively.

\vspace{-1em}
\paragraph{MS-COCO.}
The Microsoft COCO dataset~\cite{lin2014microsoft} contains 328K images with instance labels spanning 80 object classes. 
The images are also web-crawled from Flickr and include ``common'' objects like people, animals, vehicles, furniture, food and more. 
We motivate this choice of dataset because, like Open Images, it is widely used for training and benchmarking models.
We only sample images from the val2017 split (5K images) as the test split does not have ground-truth labels that are publicly available.

\subsection{CLIP variants}

We include all CLIP variants we study in~\cref{app:tab:clip-variants}, including their pre-training dataset (and size) and model checkpoint. All checkpoints are taken from {\small \href{https://github.com/mlfoundations/open_clip}{\texttt{open\_clip}}}~\cite{ilharcogabriel2021}.

\begin{table*}
    \centering
    \caption{All CLIP variants with their pre-training dataset, pre-training dataset size, and checkpoint (taken from \href{https://github.com/mlfoundations/open_clip}{\texttt{open\_clip}}~\cite{ilharcogabriel2021}). }
    \label{app:tab:clip-variants}
    \scalebox{0.93}{
    \begin{tabular}{c|cccc}
    \toprule
    \textbf{CLIP variant} & \textbf{Pre-training dataset} & \textbf{Dataset size} & \textbf{Checkpoint} \\
    \midrule
        ViT-B/16 & WIT~\cite{radford2021learning} & 400M & \texttt{openai} &  \\
        ViT-B/16 & LAION-80M~\cite{jia2021scaling} & 80M & \texttt{Data-80M\_Samples-34B\_lr-1e-3\_bs-88k} \\
        ViT-B/16 & LAION-400M~\cite{schuhmann2021laion400m} & 400M & \texttt{laion400m\_e32} \\
        ViT-B/16 & LAION-2B~\cite{schuhmann2022laion5B} & 2B & \texttt{laion2b\_s34b\_b88k} \\
        ViT-B/16 & DataComp-L~\cite{gadre2023datacomp} & 140M & \texttt{datacomp\_l\_s1b\_b8k} \\
        ViT-B/16 & CommonPool-L~\cite{gadre2023datacomp} & 1.28B & \texttt{commonpool\_l\_s1b\_b8k} \\
        ViT-B/16 & CommonPool-L (CLIP-Score filt.)~\cite{gadre2023datacomp} & 384M & \texttt{commonpool\_l\_clip\_s1b\_b8k} \\
        \midrule
        ViT-B/32 & WIT~\cite{radford2021learning} & 400M & \texttt{openai} \\
        ViT-B/32 & LAION-80M~\cite{jia2021scaling} & 80M & \texttt{Data-80M\_Samples-34B\_lr-1e-3\_bs-88k} \\
        ViT-B/32 & LAION-400M~\cite{schuhmann2021laion400m} & 400M & \texttt{laion400m\_e32} \\
        %ViT-B/32 & LAION-400M & \texttt{Model-B-32\_Data-400M\_Samples-34B\_lr-5e-4\_bs-32k} \\
        ViT-B/32 & LAION-2B~\cite{schuhmann2022laion5B} & 2B & \texttt{laion2b\_s34b\_b79k}  \\
        ViT-B/32 & DataComp-S~\cite{gadre2023datacomp} & 1.4M &  \texttt{datacomp\_s\_s13m\_b4k} \\
        ViT-B/32 & DataComp-M~\cite{gadre2023datacomp} & 14M &  \texttt{datacomp\_m\_s128m\_b4k} \\
        ViT-B/32 & CommonPool-S~\cite{gadre2023datacomp} & 12.8M & \texttt{commonpool\_s\_s13m\_b4k} \\
        ViT-B/32 & CommonPool-S (CLIP-Score filt.)~\cite{gadre2023datacomp} & 3.8M & \texttt{commonpool\_s\_clip\_s13m\_b4k} \\
        ViT-B/32 & CommonPool-M~\cite{gadre2023datacomp} & 128M & \texttt{commonpool\_m\_s128m\_b4k} \\
        ViT-B/32 & CommonPool-M (CLIP-Score filt.)~\cite{gadre2023datacomp} & 38M & \texttt{commonpool\_m\_clip\_s128m\_b4k} \\
        \midrule
        ViT-L/14 & WIT~\cite{radford2021learning} & 400M & \texttt{openai}  \\ 
        ViT-L/14 & LAION-80M~\cite{jia2021scaling} & 80M & \texttt{Data-80M\_Samples-34B\_lr-1e-3\_bs-88k} \\
        ViT-L/14 & LAION-400M~\cite{schuhmann2021laion400m} & 400M & \texttt{laion400m\_e32} \\
        %ViT-L/14 & LAION-400M & \texttt{Model-L-14\_Data-400M\_Samples-34B\_lr-1e-3\_bs-86k} \\
        ViT-L/14 & LAION-2B~\cite{schuhmann2022laion5B} & 2B & \texttt{laion2b\_s32b\_b82k}  \\
        ViT-L/14 & DataComp-XL/1B~\cite{gadre2023datacomp} & 1.4B & \texttt{datacomp\_xl\_s13b\_b90k}  \\
        ViT-L/14 & CommonPool-XL (CLIP-Score filt.)~\cite{gadre2023datacomp} & 3.8B & \texttt{commonpool\_xl\_clip\_s13b\_b90k}  \\
        \midrule
        ViT-H/14 & LAION-2B~\cite{schuhmann2022laion5B} & 2B & \texttt{laion2b\_s32b\_b79k} \\
        \midrule
        ViT-g/14 & LAION-2B~\cite{schuhmann2022laion5B} & 2B & \texttt{laion2b\_s34b\_b88k} \\
    \bottomrule
    \end{tabular}}
    \vspace{-0.8em}
\end{table*}

\subsection{Disability and exclusive disability objects}
\label{app:sec:disability-objs}

Three annotators manually categorized the 486 ORBIT objects into 55 disability objects, 42 exclusive disability objects (a subset of disability objects) and 431 non-disability objects.
Of these, 39, 30 and 310 were unique disability, exclusive disability and non-disability objects, respectively.
We include the object lists for each category below: 
\vspace{-1em}
\paragraph{Unique disability objects [39 objects]:} folded cane, solo audiobook player, orbit braille reader and notetaker, victor stream book reader, cane, white cane, digital recorder, magnifier, long cane, pen friend, braille note, dog poo, symbol cane, pocket magnifying glass, glasses, folded long guide cane, insulin pen, dictaphone, white mobility came, dog lead, retractable dog lead, braille orbit reader, victor reader stream, dogs lead, my hearing aid, water level sensor, braillepen slim braille keyboard, guide dog play cola, black mobility cane, my braille displat, visibility stick, leash, inhaler, liquid level indicator, hearing aid, guide dog harness, orbit reader 20 braille display, folded white cane, my cane

\vspace{-1em}
\paragraph{Unique exclusive disability objects [30 objects]:} folded cane, solo audiobook player, orbit braille reader and notetaker, victor stream book reader, cane, white cane, digital recorder, magnifier, long cane, pen friend, braille note, dog poo, symbol cane, pocket magnifying glass, dictaphone, folded long guide cane, white mobility came, braille orbit reader, victor reader stream, my hearing aid, water level sensor, braillepen slim braille keyboard, black mobility cane, my braille displat, visibility stick, liquid level indicator, hearing aid, orbit reader 20 braille display, folded white cane, my cane

\vspace{-1em}
\paragraph{Unique non-disability objects [310 objects]:} cushion, tred mill, apple airpods, headphones, ipod stand, wallet for bus pass cards and money, handheld police scanner, shelf unit with things, av tambourine, tea, toothbrush, door, door keys, lotion bottle, pint glass, favourite earings, prosecco, apple mobile phone, hat, tumble dryer, wall plug, risk watch, green water bottle, apple earpods, hole punch, phone stand, aspirin, tablets, garden shed, desk, knitting basket, dark glasses, headphone case, bin, chap stick, blue headphones, ottawa bus stop, fire stick remote, perfume, hair clip, pink himalayan salt, my purse, yellow marker, ipod in wallet, deodorant, mobile phone, iphone stand, apple phone charger, pencil case, one cup kettle, phone charger, adaptive dryer, skip prep, sunglasses case, eyewear case, apple headphones, front door, cranberry cream tea, backpack, keychain, 13 measuring cup, microwave, apple wireless keyboard, my tilly hat, dog toy, speaker, water bottle, my airpods, garden table, ruler, journal, stairgate, sleep mask, coffee mug, radar key, lighter, trainer shoe, toaster, vape pen, banana, house keys, winter gloves, cannabis vape battery, my tilly hat upside down, cap, small space screwdriver, dab radio, watering can, wheely bin, litter and dog waste bin, my headphones, my muse s headband, airpods, set of keys, wireless earphones, iphone in case, pink marker, scissors, blue tooth keyboard, remote control, my wraparound sunglasses, finger nail clipper, vagabond ale bottle, face mask, screwdriver, sock, front door to house, my mug, single airpod, back patio gate, earphones, 14 measuring cup, sky q remote, tv unit, lip balm, reptile green marker, coin purse, post box, watch, t-shirts, bus stop sign, buckleys, ladies purse, iphone air pods, recycling bin, black bin, key, black small wallet, table fan, exercise bench, keyboard, hand gel, purse, vase with flowers, white came, house door, wallet, reading glasses, orange skullcap, baked bean tin, migenta marker, my purple mask, condom box, mediterranean sea salt, my work backpack, personal mug, bottle opener, my slate, clickr, measuring spoon, rice, mug, iphone 6, presentation remote, secateurs, ps4 controller, remote tv, necklace, wardrobe, aspirin vs tylenol, mouse, small screwdriver, socks, eye drops, mustard, hand saw, lipstick, bose wireless headphones, hair brush, hairbrush, pinesol cleaner, memory stick, glasses case, knitting needle, pepper shaker, cup again, bone conducting headset, fridge, usb stick, compact disc, work phone, wine glass, my front door, work bag, headband, airpod pro, walletv, my laptop, money pouch, remote, jd whisky bottle, paperclips, pex plumbers pliers, samsung tv remote control, my airpod pros case, portable keyboard, money clip, flat screen television, clear nail varnish, usb c dongle, amazon remote control, digital dab radio, 1 cup, measuring cup, tissue box, baseball cap, earpods, gloves, p939411 white cane, smarttv, skipping rope, back door, i d wallet, bluetooth keyboard, sunglasses, headset, my pill dosette, fridge freezer indicator, usb, apple pencil, black strappy vest, my apple watch, cell phone, apple wath, airpods pro charging case, slippers, dog streetball, corkscrew, airpod case, veg peeler, local post box, brown leather bracelet, pill bottle, my wallet, medication, mayonnaise jar, sofa, bottle, virgin remote control, money, slipper, fish food, styrofoam cup, blue facemask, i phone 11 pro, my keyboard, ipad, nobile phone stand, glasses cleaning wipe, bottle of alcoholic drink, cooker, tv remote, front door keys, tweezers, shed door, kettle, alcohol wipe, make up, battery drill, spanner, apple tv remote, bag, phone case, mini bluetooth keyboard, stylus, shoulder bag, comb, my keys, mirror, my clock, eye glasses, nike trainers, my water bottle, garden wall, sharp knife, my shoes, back pack, grinder, 12 measuring cup, iphone, phone, covid mask, mountain dew can, wheelie bin, car, headphone, keys, large sewing needle, miter saw, apple watch, chicken instant noodles, tv remote control, adaptive tennis ball, embroidery thread cone, washing basket, wrist watch, lime green marker, glass, boot, bed, bose earpods, television remote control, dining table setup, toddler cup, tape measure, adaptive washing machine, pop bottle, electric sanding disc, washing machine, my sennheiser pxc 350-2, ladies silver bracelet

\subsection{Colors and materials}
\label{app:sec:colors-materials}

Three annotators manually annotated the ORBIT validation and test objects (208 objects) with their color and material.
In most cases, each object was labeled with one color and one material, but in some cases up to two labels were selected (\eg a water bottle with a plastic body and metal lid was assigned ``plastic metal'' as its material).
The labels were iterated until all three annotators agreed.
All colors and materials were selected from the following lists:

\vspace{-1em}
\paragraph{Colors [20 colors]:} red, silver, yellow, grey, dark, pink, multicolour, purple, white, beige, burgundy, maroon, blue, green, black, gold, brown, light, transparent, orange

\vspace{-1em}
\paragraph{Materials [23 materials]:} rubber, crystal, cardboard, denim, material, styrofoam, stone, glass, foam, cloth, leather, ceramic, plastic, wood, paper, embroidered, wooden, suede, canvas, patterned, metal, cotton, lacquered

\subsection{Textual analysis of LAION-400M, LAION-2B and DataComp-1B}
\label{app:sec:dataset-analysis}

Our aim is to quantify the prevalence of disability content in large-scale datasets used to pre-train \glspl{lmm} -- specifically, LAION-400M~\cite{schuhmann2021laion400m}, LAION-2B~\cite{schuhmann2022laion5B} and DataComp-1B (or XL)~\cite{gadre2023datacomp}.
To do this, we first extract all visual concepts from the captions of each dataset (see~\cref{app:alg:extract-nps}).
We define a visual concept as a noun phrase that contains a physical object (\eg ``park bench'').
We consider a noun phrase as a phrase that contain a common noun and optional adjectives (\eg ``green park bench'').
We consider the common noun to be a physical object if it traverses the ``entity”, ``physical\_entity”, and ``object" hypernyms and then either the ``artifact", ``whole", ``part", or ``living\_thing" hypernym in the WordNet tree hierarchy~\cite{miller1995wordnet} (see~\cref{app:alg:is-physical-object}).
\begin{algorithm*}
\caption{Pseudocode for \texttt{\small extract\_noun\_phrases(captions: List[str]) -> List[str]}}
\label{app:alg:extract-nps}
\begin{algorithmic}[1]
\State \texttt{\small regex\_pattern = r"""NP: {<DT|PRP\$\textbackslash\$>?<JJ>*<NN|NNS>}""" \textcolor{blue}{\# a noun phrase (NP) contains a singular or plural noun (NN/NNS) which may be prefixed by an article (DT)/possessive pronoun (PRP) and/or adjectives (JJ)}}
\State \texttt{\small chunker = nltk.RegexpParser(regex\_pattern)}
\State \texttt{\small noun\_phrases = []}
\For {\texttt{\small caption in captions}}
\State \texttt{\small tokens = nltk.word\_tokenize(caption.lower()) \textcolor{blue}{\# tokenize}}
\State \texttt{\small pos\_tags = nltk.pos\_tag(tokens) \textcolor{blue}{\# extract parts of speech }}
\State \texttt{\small np\_tree = chunker(pos\_tags) \textcolor{blue}{\# extract noun phrase tree}}
\State \texttt{\small noun = extract\_noun(np\_tree) \textcolor{blue}{\# extract noun (NN or NNS) from tree}}
    \If {\texttt{\small is\_physical\_object(noun)}}
        \State \texttt{\small cleaned\_np = clean\_np(np\_tree) \textcolor{blue}{\# remove DT/PRP and singularize noun} }
        \State \texttt{\small noun\_phrases.extend(cleaned\_np)}
    \EndIf
\EndFor
\State \texttt{\small return noun\_phrases}
\end{algorithmic}
\end{algorithm*}
\begin{algorithm*}
\caption{Pseudocode for \texttt{\small is\_physical\_object(word: str) -> bool:}}
\label{app:alg:is-physical-object}
\begin{algorithmic}[1]
\State \texttt{\small synsets = wordnet.synsets(word, "n")  \textcolor{blue}{\# get WordNet noun synsets}}
\For{\texttt{\small synset in synsets}}
    \State \texttt{\small paths = synset.hypernym\_paths() \textcolor{blue}{\# get hypernym paths}}
    \For{\texttt{\small path in paths}}
        \State \textcolor{blue}{\texttt{\small \# path is a list e.g. [Synset("entity.n.01"), ..., Synset("bench.n.01")]}}
        \State \texttt{\small if (path contains "entity.n" AND "physical\_entity.n" AND "object.n" \textbackslash }
        \State \quad \texttt{\small AND ("artifact.n" OR "whole.n" OR "part.n" OR "living\_thing.n")):}
        \State \quad \texttt{\small return True}
    \EndFor
\EndFor
\State \texttt{\small return False}
\end{algorithmic}
\end{algorithm*}
In~\cref{app:tab:noun-phrases} we report the top ten noun phrases extracted from LAION-400M, LAION-1B and DataComp-1B.
We see that many of these are shared across all three datasets, including ``image'', ``photo'', ``man'', and ``woman''.
\begin{table*}[h]
\centering
\caption{Top 10 noun phrases extracted from the captions of the LAION-400M~\cite{schuhmann2021laion400m}, LAION-2B~\cite{schuhmann2022laion5B} and DataComp-1B~\cite{gadre2023datacomp} datasets. See extraction protocol in~\cref{app:sec:dataset-analysis}.}
\scalebox{0.9}{
\begin{tabular}{cc|cc|cc}
\toprule
\multicolumn{2}{c|}{\textbf{LAION-400M}} & \multicolumn{2}{c|}{\textbf{LAION-2B}} &  \multicolumn{2}{c}{\textbf{DataComp-1B}}\\
%\hline{2-6}
\textbf{Noun phrase}  & \textbf{Occurrence count} & \textbf{Noun phrase} & \textbf{Occurrence count} & \textbf{Noun phrase} & \textbf{Occurrence count} \\
\midrule
image & 8,930,057 & image & 69,739,546 & image & 35,784,938  \\
photo & 7,559,650 & photo & 55,970,047 & photo & 28,612,087  \\
vector & 4,523,146 & vector & 29,203,691 & vector & 14,157,166  \\
man & 3,458,074 & stock & 22,209,987 & stock & 13,062,936  \\
design & 3,052,442 & man & 21,261,979 & background & 10,829,331  \\
background & 2,760,736 & background & 20,873,562 & design & 10,473,440  \\
woman & 2,513,375 & picture & 19,322,717 & home & 8,731,799  \\
home & 2,446,557 & design & 18,335,833 & picture & 8,523,946  \\
stock & 2,236,958 & home & 17,747,460 & man & 8,453,533  \\
picture & 2,235,123 & woman & 17,001,793 & view & 7,595,762 \\
\bottomrule
\end{tabular}}
\label{app:tab:noun-phrases}
\end{table*}

\vspace{-1em}
\subsubsection{Prevalence of disability vs non-disability objects}
In~\cref{sec:image-content:dataset-analysis-dis-vs-non-dis} of the main paper, we quantify how often disability and non-disability objects occur in the extracted visual concepts.
We use the ORBIT objects as a seed set (see lists in~\cref{app:sec:disability-objs}).
Three annotators first grouped the object labels into clusters based on object similarity (\eg all guide canes, all spectacles).
Two synonyms were then assigned per cluster to account for different ways objects can be described.
Early results showed that the disability clusters' synonyms occurred extremely rarely in the visual concepts, so we broadened this to 5-16 synonyms per cluster.
The 222 and 312 synonyms for the disability and non-disability clusters are provided in~\cref{app:tab:dataset-analysis:synonyms-dis-objs} and~\cref{app:tab:dataset-analysis:synonyms-non-dis}, respectively.

We then count how many times each of these synonyms appears in the extracted visual concepts (see counts in~\cref{app:tab:dataset-analysis:synonyms-dis-objs,app:tab:dataset-analysis:synonyms-non-dis}).
We do this using direct string matching allowing for partial matches (\eg ``braille note taker'' is marked as present in the visual concept ``cheap braille note taker'').
Before matching, we lower-case and remove punctuation from all synonyms and visual concepts following typical VQA practices  (see \texttt{\small processPunctuation} function in the \href{https://github.com/GT-Vision-Lab/VQA/blob/master/PythonEvaluationTools/vqaEvaluation/vqaEval.py}{GT-Vision-Lab/VQA repo}).
For synonyms that contain multiple words, we also allow for multiple spellings (\eg tread mill and treadmill).

\begin{table*}
\renewcommand{\arraystretch}{0.5}% Tighter
\centering
\caption{Disability object clusters, their synonyms, and their prevalence in the LAION-400M (L400M), LAION-2B (L2B) and DataComp-1B (DC1B) datasets. Numbers reported are the total number of times each cluster's synonyms appeared in the dataset's extracted visual concepts (total visual concepts -- LAION-400M: 384,468,921; LAION-2B: 2,737,763,447; and DataComp-1B: 1,342,369,058).}
\scalebox{0.8}{
\begin{tabular}{L{2.2cm}|m{14cm}|R{1cm}|R{1cm}|R{1cm}}
\hline
\toprule
\textbf{Object cluster} & \textbf{Synonyms} & \textbf{L400M} & \textbf{L2B} & \textbf{DC1B} \\
\hline
braille readers & braille note taker, braille reader, braille display, braille notetaker, braille tablet, braille computer, braille keyboard, orbit reader, braillepen slim braille keyboard, braillepen slim keyboard & 4  & 8  & 6  \\\hline
dictaphones & dictaphone, digital recorder, voice recorder, dictation machine, audio recorder, voice recording device, dictation recorder, audio dictation device, voice transcription device, handheld recorder & 40  & 185  & 168  \\\hline
digital book readers & digital book player, digital book reader, victor stream, victor reader stream, talking book, humanware reader, solo audiobook player, audiobook player & 0  & 2  & 1  \\\hline
dog leads & dog lead, dogs lead, dog leash, dogs leash, leash, dog tether, dogs tether & 434  & 1,663  & 1,402  \\\hline
dog poo & dog poo, dog poop, dog waste, dog scat, dog dung, canine faeces, canine feces, canine faeces & 3  & 11  & 9  \\\hline
glasses & glass, sight glass, spectacle, eyeglass, reading glass, prescription glass, optical glass, corrective lens, bi focal, eyewear, frame, multi focal, optical, vision aid, spec & 17,620 & 68,169  & 46,259 \\\hline
guide canes & guide cane, symbol cane, mobility cane, long cane, white cane, blind cane, white mobility cane, vision cane, assistive cane, visibility stick & 21  & 79  & 50  \\\hline
hearing aids & hearing aid, hearing device, hearing amplifier, assistive listening device, hearing implant, cochlear implant, audio prosthesis, auditory prosthesis & 23  & 122  & 77  \\\hline
inhalers & inhaler, asthma pump, asthma puffer, aerosol inhaler, inhalant delivery system & 81  & 282  & 315  \\\hline
insulin pens & insulin pen, insulin injector, insulin delivery pen, insulin auto-injector, insulin syringe pen, insulin dispenser, insulin delivery system, insulin applicator, insulin dosing pen, diabetes pen & 2  & 4  & 4  \\\hline
liquid level sensors & liquid level sensor, liquid level indicator, liquid level detector, liquid level gauge, water level sensor, water level indicator, water level detector, water level gauge & 1  & 5  & 8  \\\hline
magnifiers & magnifier, magnifying glass, magnification aid, magnifying lens & 89  & 390  & 362  \\\hline
audio labelers & penfriend, pen friend, audio labeller, audio labelling device, audio labelling pen, audio labelling tool, voice labeller, voice labelling pen, voice labelling device, voice labelling tool, speech-enabled labeller, speech-enabled labelling device, speech-enabled labelling pen, speech-enabled labelling tool, talking label maker, speech-based label printer & 8  & 19  & 11  \\
\bottomrule
\end{tabular}}
\label{app:tab:dataset-analysis:synonyms-dis-objs}
\vspace{-0.5em}
\end{table*}

\begin{table*}
\renewcommand{\arraystretch}{0.5}% Tighter
\centering
\caption{Non-disability object clusters, their synonyms, and their prevalence in the LAION-400M (L400M), LAION-2B (L2B) and DataComp-1B (DC1B) datasets. Numbers reported are the total number of times each cluster's synonyms appeared in the dataset's extracted visual concepts (total visual concepts -- LAION-400M: 384,468,921; LAION-2B: 2,737,763,447; and DataComp-1B: 1,342,369,058).}
\scalebox{0.7}{
\begin{tabular}{L{2.4cm}m{5cm}|R{0.9cm}|R{0.9cm}|R{0.9cm}|L{2.6cm}m{4.7cm}|R{0.9cm}|R{0.9cm}|R{0.9cm}}
\toprule
\textbf{Object cluster} & \textbf{Synonyms} & \textbf{L400M} & \textbf{L2B} & \textbf{DC1B} & \textbf{Object cluster} & \textbf{Synonyms} & \textbf{L400M} & \textbf{L2B} & \textbf{DC1B}  \\
\midrule
airpods & airpod, ear phone & 655  & 2,077  & 1,763  & make-up & make-up, make up & 6,202  & 23,424  & 15,361  \\\hline
airpods cases & airpods case, airpods pro case & 0  & 1  & 2  & markers & marker, felt-tip pen & 1,254  & 5,525  & 3,634  \\\hline
alcohol wipes & alcohol wipe, alcohol pad & 1  & 1  & 0  & measuring spoons & measuring spoon, measuring cup & 10  & 50  & 55  \\\hline
bags & bag, backpack & 15,250 & 52,351  & 34,650  & medications & medication, pill & 689  & 2,477  & 2,415  \\\hline
balls & ball, dog toy & 6,540  & 23,289  & 14,888  & mirrors & mirror, looking glass & 4,755  & 18,788  & 13,327  \\\hline
bananas & banana, fruit & 7,631  & 25,879  & 21,884  & mice & bluetooth mouse, wireless mouse & 3  & 13  & 16  \\\hline
baskets & basket, crate & 3,375  & 12,380  & 8,821  & mugs & mug, cup & 10,360  & 39,145  & 30,334  \\\hline
beds & bed, mattress & 8,087  & 35,714  & 23,575  & nail clippers & nail clipper, tweezers & 10  & 48  & 76  \\\hline
beers & beer, alcohol & 458  & 2,145  & 1,564  & nail polishes & nail polish, nail varnish & 916  & 2,918  & 1,787  \\\hline
bins & bin, trash can & 1,265  & 4,526  & 3,370  & needles & needle, pin & 7,531  & 27,235  & 23,560  \\\hline
bottles & bottle, thermos & 6,113  & 21,684  & 20,062  & paper clips & paper clip, paper fastener & 106  & 386  & 366  \\\hline
bottle openers & bottle opener, cork screw & 127  & 515  & 683  & peelers & peeler, scraper & 294  & 1,085  & 1,259  \\\hline
bracelets & bracelet, necklace & 18,438 & 63,242  & 61,734 & pencil cases & pencil case, pen case & 33  & 161  & 152  \\\hline
brushes & brush, comb & 3,866  & 13,624  & 11,820  & phones & phone, iphone & 5,217  & 19,911  & 11,396 \\\hline
bus stops & bus stop, bus station & 36  & 163  & 103  & phone chargers & phone charger, charging cable & 2  & 13  & 17  \\\hline
cans & can, tin & 2,648  & 10,906  & 7,629  & phone stands & phone stand, ipad stand & 9  & 32  & 17  \\\hline
cars & car, vehicle & 22,867 & 84,568 & 57,283 & plugs & plug, socket & 3,340  & 12,519  & 10,972  \\\hline
CDs & compact disc, cd & 9,675  & 31,511  & 14,609  & police scanners & police scanner, radio scanner & 1  & 2  & 2  \\\hline
cleaners & cleaner, surface spray & 956  & 3,767  & 2,850  & pops & pop, soda & 5,795  & 20,591  & 13,506  \\\hline
clocks & clock, timekeeper & 2,783  & 9,560  & 7,310  & post boxes & post box, mail box & 608  & 2,010  & 1,625  \\\hline
condom boxes & condom box, durex box & 0  & 1  & 0  & purses & purse, wallet & 6,221  & 21,201  & 15,256  \\\hline
cookers & cooker, air fryer & 654  & 2,612  & 2,718  & radios & radio, receiver & 4,638  & 15,645  & 12,138  \\\hline
cushions & cushion, pillow & 9,081  & 33,694  & 24,195  & rice & rice, noodle & 4,725  & 16,489  & 13,833  \\\hline
deodorants & deodorant, perfume & 1,564  & 5,875  & 5,556  & rulers & ruler, tape measure & 772  & 2,963  & 2,118  \\\hline
dog waste bins & dog waste bin, dog waste container & 0  & 0  & 0  & sauces & mustard, mayonnaise & 1,050  & 3,877  & 2,941  \\\hline
doors & door, entrance & 15,627 & 54,395  & 44,418 & scissors & secateur, scissor & 85  & 284  & 268  \\\hline
drills & drill, power tool & 1,055  & 4,115  & 3,082  & screwdrivers & screwdriver, spanner & 383  & 1,466  & 2,109  \\\hline
electric saws & electric saw, chain saw & 368  & 1,340  & 1,007  & sheds & shed, tool shed & 786  & 3,112  & 2,270  \\\hline
eye drops & eye drop, eye gel & 6  & 35  & 18  & shoes & shoe, sneaker & 11,078  & 43,179  & 21,112  \\\hline
face masks & face mask, face covering & 43  & 255  & 181  & skipping ropes & skipping rope, jump rope & 7  & 19  & 17  \\\hline
fans & fan, air cooler & 7,638  & 24,013  & 15,585  & sleep masks & sleep mask, eye mask & 44  & 158  & 144  \\\hline
fish foods & fish food, fish flake & 4  & 15  & 12  & socks & sock, sockwear & 2,959  & 8,765  & 6,443  \\\hline
fridges & fridge, freezer & 1,784  & 5,516  & 4,926  & sofas & sofa, couch & 11,588  & 47,305  & 31,437  \\\hline
game controllers & wireless controller, game controller & 5  & 29  & 36  & spices & salt, pepper & 2,814  & 9,365  & 9,068  \\\hline
gates & gate, gateway & 3,133  & 11,797  & 8,287  & styluses & apple pen, stylus & 151  & 688  & 681  \\\hline
glasses & glass, tumbler & 5,022  & 19,732  & 13,595  & sunglasses & sunglass, shade & 6,964  & 24,927  & 16,626  \\\hline
glasses cases & glasses case, sunglasses case & 2  & 5  & 4  & tables & desk, table & 23,387 & 104,138 & 69,152 \\\hline
glasses cleaners & glasses cleaner, lens wipe & 0  & 0  & 0  & tambourines & tambourine, tamborine & 89  & 362  & 375  \\\hline
gloves & glove, mitten & 3,866  & 13,548  & 8,228  & tea & tea, teabag & 7,739  & 25,126  & 20,249  \\\hline
grinders & grinder, food processor & 421  & 1,645  & 1,525  & thread cones & thread cone, thread spool & 15  & 49  & 51  \\\hline
hair clips & hair clip, headband & 1,546  & 5,115  & 4,381  & tissue boxes & tissue box, kleenex & 12  & 29  & 26  \\\hline
hand sanitizers & hand sanitizer, hand santiser & 0  & 10  & 5  & toothbrushes & toothbrush, dental brush  & 537  & 2,329  & 2,443  \\\hline
hand saws & hand saw, hack saw & 46  & 206  & 245  & tread mills & tread mill, running machine & 263  & 953  & 627  \\\hline
hats & hat, cap & 10,100  & 34,840  & 24,185  & t-shirts & t-shirt, tee & 30,770 & 100,369 & 71,408 \\\hline
headphones & headphone, headset & 2,241  & 7,843  & 6,396  & TVs & tv, television & 15,350 & 53,606  & 35,370  \\\hline
headphone cases & headphone case, headphones case & 0  & 1  & 1  & TV remotes & tv remote, remote control & 203  & 866  & 658  \\\hline
hole punches & hole punch, paper punch & 47  & 174  & 142  & USB sticks & usb stick, flash drive & 221  & 651  & 690  \\\hline
iPads & ipad, tablet & 6,365  & 21,157  & 14,534  & vapes & vape, e-cigarette & 160  & 540  & 469  \\\hline
journals & journal, notebook & 6,201  & 24,084  & 15,205  & vases & vase, jug & 4,853  & 16,954  & 16,971  \\\hline
kettles & kettle, toaster & 1,128  & 4,879  & 4,348  & walls & wall, fence & 14,538 & 60,387  & 39,221  \\\hline
keys & key, key chain & 8,770  & 31,370  & 22,409  & wardrobes & wardrobe, cupboard & 2,101  & 8,193  & 5,926  \\\hline
keyboards & keyboard, keypad & 2,491  & 8,639  & 9,582  & washing machines & washing machine, dryer & 767  & 3,040  & 2,194  \\\hline
knives & knife, blade & 4,012  & 15,521  & 17,972  & watches & watch, smart watch & 9,651  & 36,043  & 24,128  \\\hline
laptops & laptop, chromebook & 7,522  & 24,913  & 17,948  & watering cans & watering can, water can & 14  & 110  & 66  \\\hline
lipsticks & lipstick, lip balm & 1,453  & 5,213  & 4,635  & weight benches & weight bench, gym bench & 10  & 32  & 33  \\\hline
\end{tabular}}
\label{app:tab:dataset-analysis:synonyms-non-dis}
\end{table*}

\vspace{-1.1em}
\subsubsection{Prevalence of colors vs materials}

In~\cref{sec:text-content:dataset-analysis:material} of the main paper, we quantify how often colors and materials occur in each dataset.
We use the color and materials in~\cref{app:sec:colors-materials} and directly count their frequency (via partial string matching, as above) in each dataset's extracted visual concepts (see counts in~\cref{app:tab:dataset-analysis:materials-vs-colors}).
\begin{table}[]
\centering
\caption{Colors/materials and their prevalence in the LAION-400M (L400M), LAION-2B (L2B) and DataComp-1B (DC1B) datasets. Numbers reported are the total number of times each color/material appeared in the dataset's extracted visual concepts (total visual concepts -- LAION-400M: 384,468,921; LAION-2B: 2,737,763,447; and DataComp-1B: 1,342,369,058).}
\label{app:tab:dataset-analysis:materials-vs-colors}
\scalebox{0.9}{
\begin{tabular}{c|c|r|r|r}
\toprule
& & \textbf{L400M} & \textbf{L2B} & \textbf{DC1B} \\
\hline
\multirow{20}{*}{\rotatebox[origin=c]{90}{\textbf{Colors}}} & beige & 2,462 & 8,857 & 5,265 \\
& black & 87,366 & 323,959 & 207,730 \\
& blue & 53,947 & 193,504 & 131,665 \\
& brown & 21,904 & 84,994 & 55,553 \\
& burgundy & 1,261 & 4,127 & 2,446 \\
& dark & 10,888 & 36,735 & 25,818 \\
& gold & 5,882 & 19,564 & 12,894 \\
& green & 38,876 & 136,448 & 92,922 \\
& grey & 12,816 & 47,470 & 28,847 \\
& light & 31,413 & 120,232 & 92,203 \\
& maroon & 613 & 2,367 & 1,495 \\
& multicolour & 278 & 652 & 404 \\
& orange & 10,138 & 30,733 & 23,262 \\
& pink & 14,925 & 60,226 & 35,658 \\
& purple & 10,985 & 37,673 & 25,270 \\
& red & 45,267 & 165,576 & 104,289 \\
& silver & 8,789 & 33,614 & 27,766 \\
& transparent & 2,513 & 10,098 & 6,807 \\
& white & 94,014 & 366,224 & 236,456 \\
& yellow & 20,723 & 73,049 & 49,121 \\
\midrule
\multirow{23}{*}{\rotatebox[origin=c]{90}{\textbf{Materials}}} & canvas & 897 & 3,254 & 1,920 \\
& cardboard & 451 & 1,546 & 1,154 \\
& ceramic & 12,808 & 48,649 & 43,339 \\
& cloth & 3,498 & 13,528 & 8,968 \\
& cotton & 12,619 & 47,188 & 28,071 \\
& crystal & 12,663 & 46,968 & 36,329 \\
& denim & 5,130 & 15,265 & 8,720 \\
& embroidered & 2,090 & 8,026 & 3,904 \\
& foam & 675 & 2,234 & 1,894 \\
& glass & 3,264 & 11,420 & 7,889 \\
& lacquered & 351 & 1,679 & 1,077 \\
& leather & 1,214 & 4,234 & 2,505 \\
& material & 9,287 & 44,848 & 25,692 \\
& metal & 3,183 & 11,780 & 9,198 \\
& paper & 14,615 & 58,584 & 40,500 \\
& patterned & 1,057 & 3,841 & 1,947 \\
& plastic & 4,136 & 14,748 & 10,997 \\
& rubber & 4,620 & 18,765 & 12,234 \\
& stone & 7,262 & 28,321 & 20,570 \\
& styrofoam & 83 & 294 & 257 \\
& suede & 2,787 & 10,929 & 5,197 \\
& wood & 10,947 & 43,897 & 30,791 \\
& wooden & 18,239 & 73,016 & 51,445 \\
\bottomrule
\end{tabular}}
\end{table}

\section{Extended results}

\subsection{\gls{blv} versus web-crawled data}

\begin{figure}
    \centering
    \includegraphics[width=0.95\linewidth]{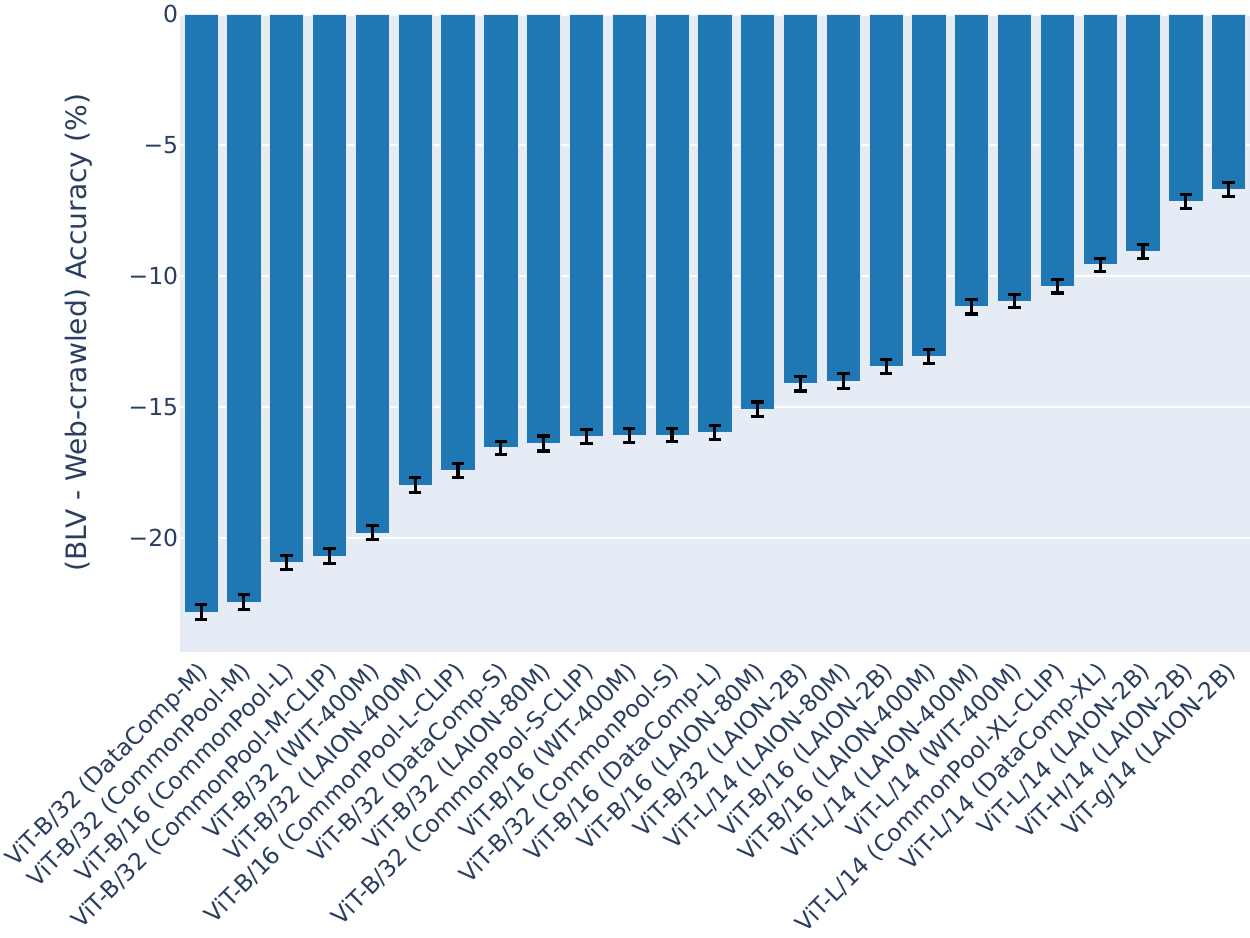}
    \caption{\textbf{No CLIP variant achieves parity between \gls{blv} and web-crawled datasets on a standarized zero-shot image classification task (see~\cref{sec:prelims:zero-shot}).} Each bar represents the variant's delta in average accuracy (with 95\% c.i.) between all images sampled from \gls{blv} versus web-crawled datasets. }
    \label{app:fig:blv-vs-webcrawled-per-model}
    \vspace{-1em}
\end{figure}

We extend~\cref{fig:blv-vs-webcrawled} in the main paper, with the delta in average accuracy between \gls{blv} and web-crawled datasets, reported per CLIP variant in~\cref{app:fig:blv-vs-webcrawled-per-model}.
We see that no model achieves parity, with smaller architectures (\eg ViT-B/32) pre-trained on smaller datasets (\eg DataComp-M, CommonPool-M) generally having a larger delta than larger architectures (\eg ViT-g/14, ViT-H/14) pre-trained on larger datasets (\eg LAION-2B).

\subsection{Robustness to image content from \gls{blv} users}

\subsubsection{Disability objects are less well recognized than
non-disability objects}

\begin{figure}
    \centering
    \begin{subfigure}{0.48\textwidth}
        \centering
        \includegraphics[width=\linewidth]{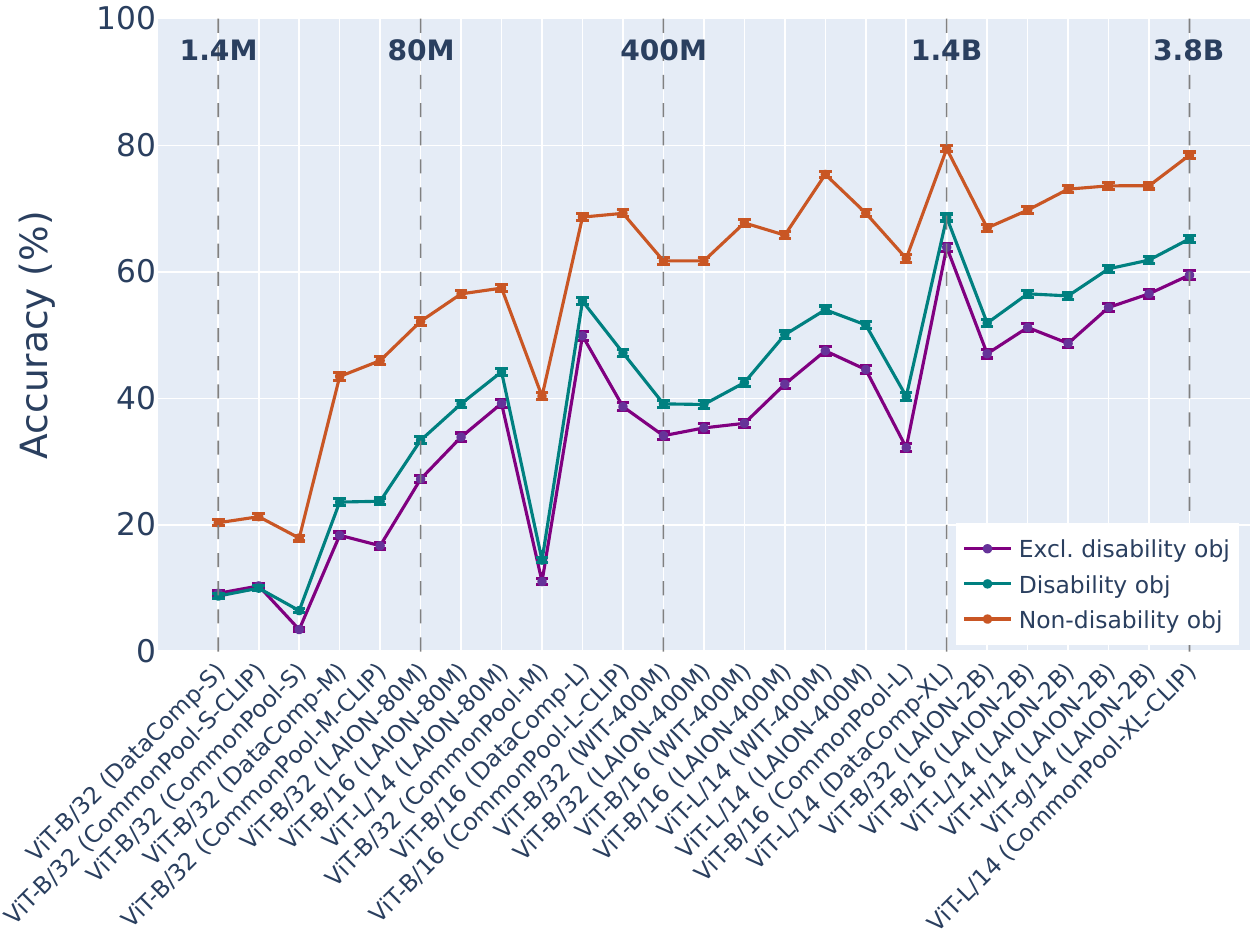}
        \caption{ORBIT Clean}
        \label{app:fig:orbit-clean-dis-vs-non-dis-by-pretraining-size}
    \end{subfigure}
    \hfill
    \begin{subfigure}{0.48\textwidth}
        \centering
        \includegraphics[width=\linewidth]{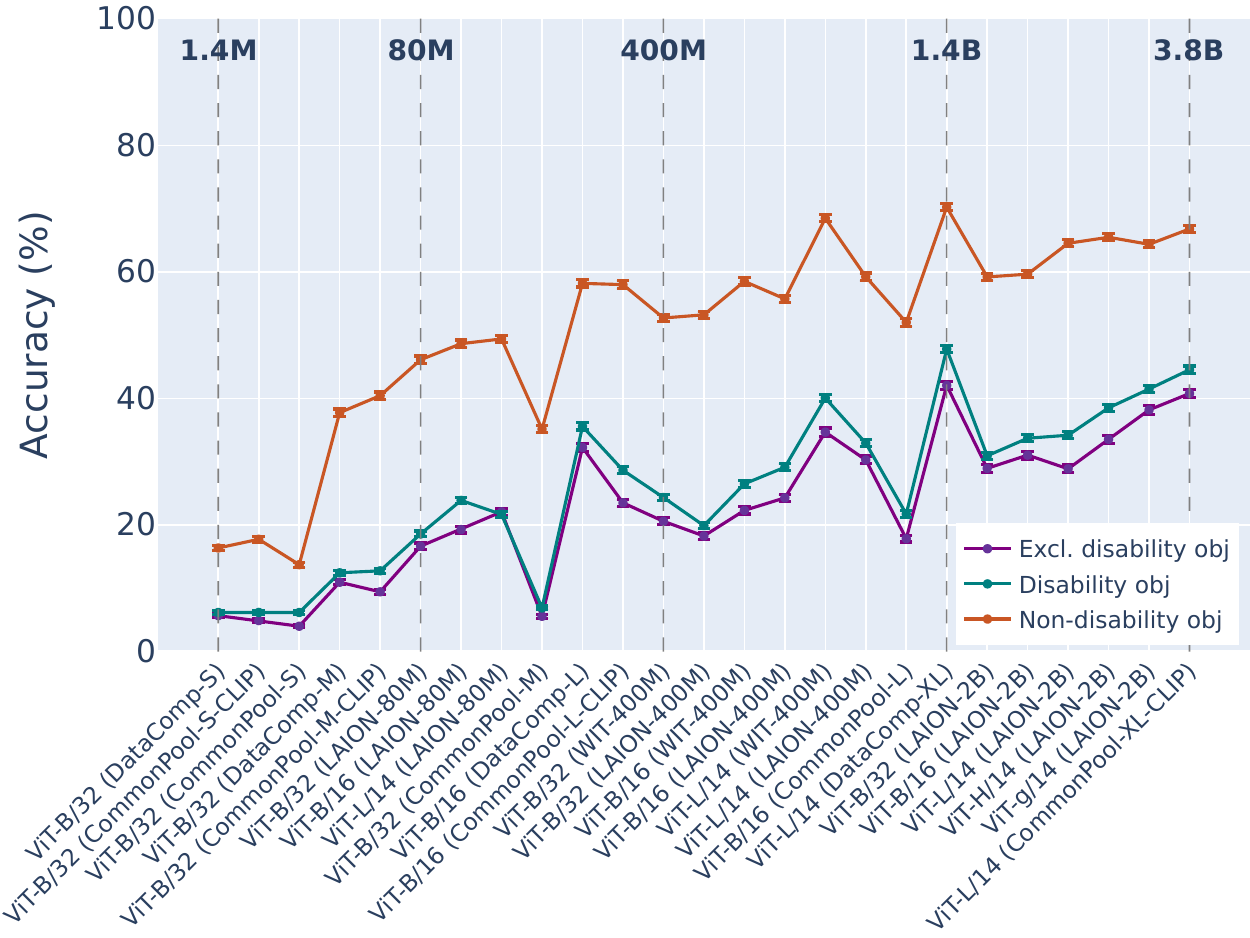}
        \caption{ORBIT Clutter}
        \label{app:fig:orbit-clutter-dis-vs-non-dis-by-pretraining-size}
    \end{subfigure}
    \caption{\textbf{CLIP's difference in accuracy between disability and non-disability objects remains largely constant as its pre-training dataset increases}. Zero-shot accuracy is averaged (with 95\% c.i.) over images from ORBIT Clean/Clutter of each object type. Experimental details in~\cref{sec:image-content:dis-vs-non-dis}.}
    \label{app:fig:orbit-clean-clutter-dis-vs-non-dis-by-pretraining-size}
    \vspace{-1em}
\end{figure}

\begin{figure}
    \centering
    \begin{subfigure}{0.48\textwidth}
        \centering
        \includegraphics[width=\linewidth]{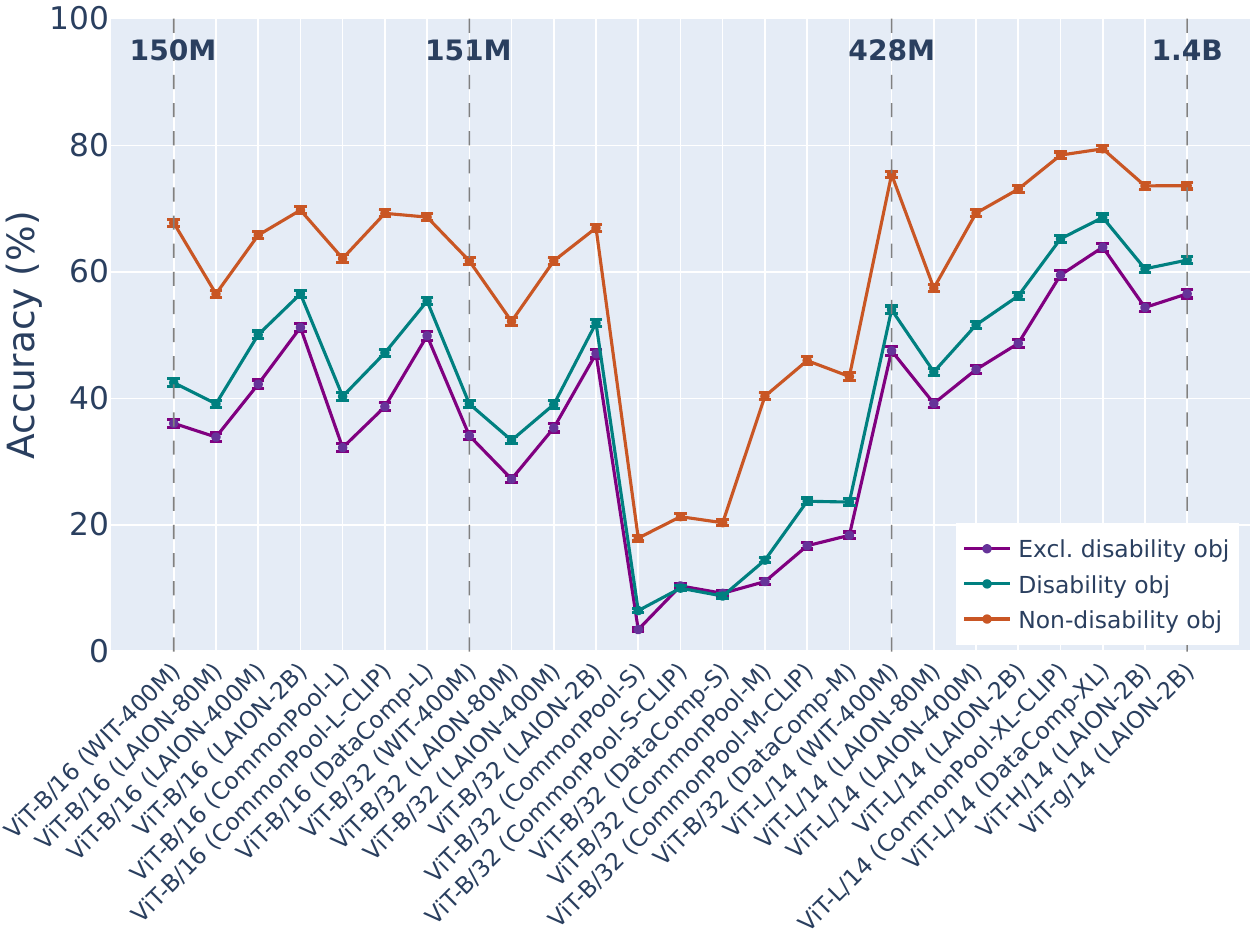}
        \caption{ORBIT Clean}
        \label{app:fig:orbit-clean-dis-vs-non-dis-by-model-size}
    \end{subfigure}
    \hfill
    \begin{subfigure}{0.48\textwidth}
        \centering
        \includegraphics[width=\linewidth]{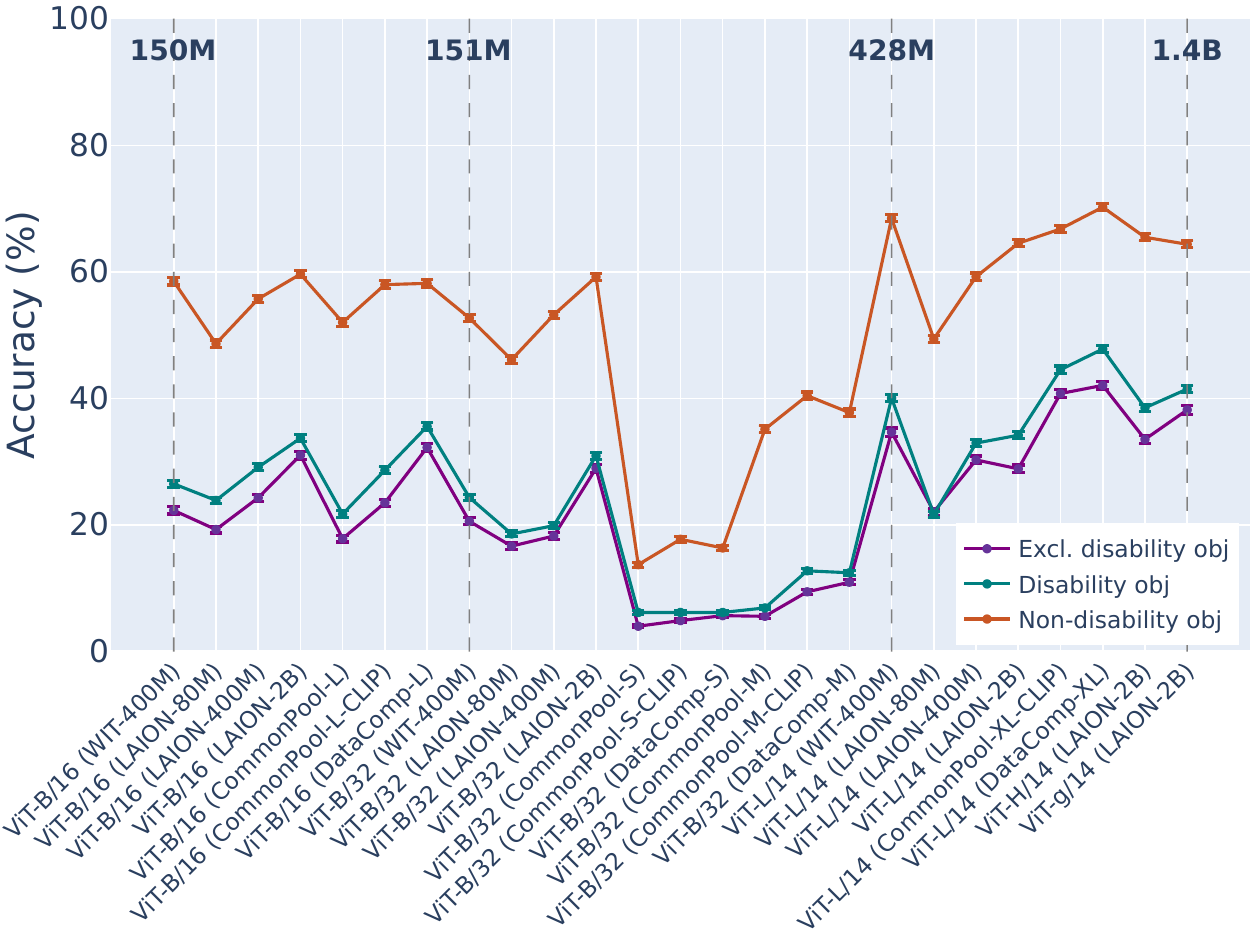}
        \caption{ORBIT Clutter}
        \label{app:fig:orbit-clutter-dis-vs-non-dis-by-model-size}
    \end{subfigure}
    \caption{\textbf{CLIP's difference in accuracy between disability and non-disability objects remains largely constant as its architecture size increases}. Zero-shot accuracy is averaged (with 95\% c.i.) over images from ORBIT Clean/Clutter of each object type. Experimental details in~\cref{sec:image-content:dis-vs-non-dis}.}
    \label{app:fig:orbit-clean-clutter--dis-vs-non-dis-by-model-size}
\end{figure}

\cref{app:fig:orbit-clean-clutter-dis-vs-non-dis-by-pretraining-size} shows the difference in average accuracy for disability, exclusive disability and non-disability objects for each CLIP variant, ordered by pre-training dataset size, on the ORBIT Clean (\cref{app:fig:orbit-clean-dis-vs-non-dis-by-pretraining-size}) and ORBIT Clutter (\cref{app:fig:orbit-clutter-dis-vs-non-dis-by-pretraining-size}) datasets.
\cref{app:fig:orbit-clean-dis-vs-non-dis-by-model-size} and~\cref{app:fig:orbit-clutter-dis-vs-non-dis-by-model-size} show the same, but with the CLIP variants ordered by architecture size.
From these figures, we see that the accuracy difference between disability/exclusive disability objects and non-disability objects remains largely constant regardless of test dataset, pre-training dataset size, and architecture size.

\vspace{-1em}
\subsubsection{A few-shot approach can \emph{sometimes} reduce the disability and non-disability accuracy gap}

\begin{figure}
    \centering
    \begin{subfigure}{0.45\textwidth}
        \centering
        \includegraphics[width=\linewidth]{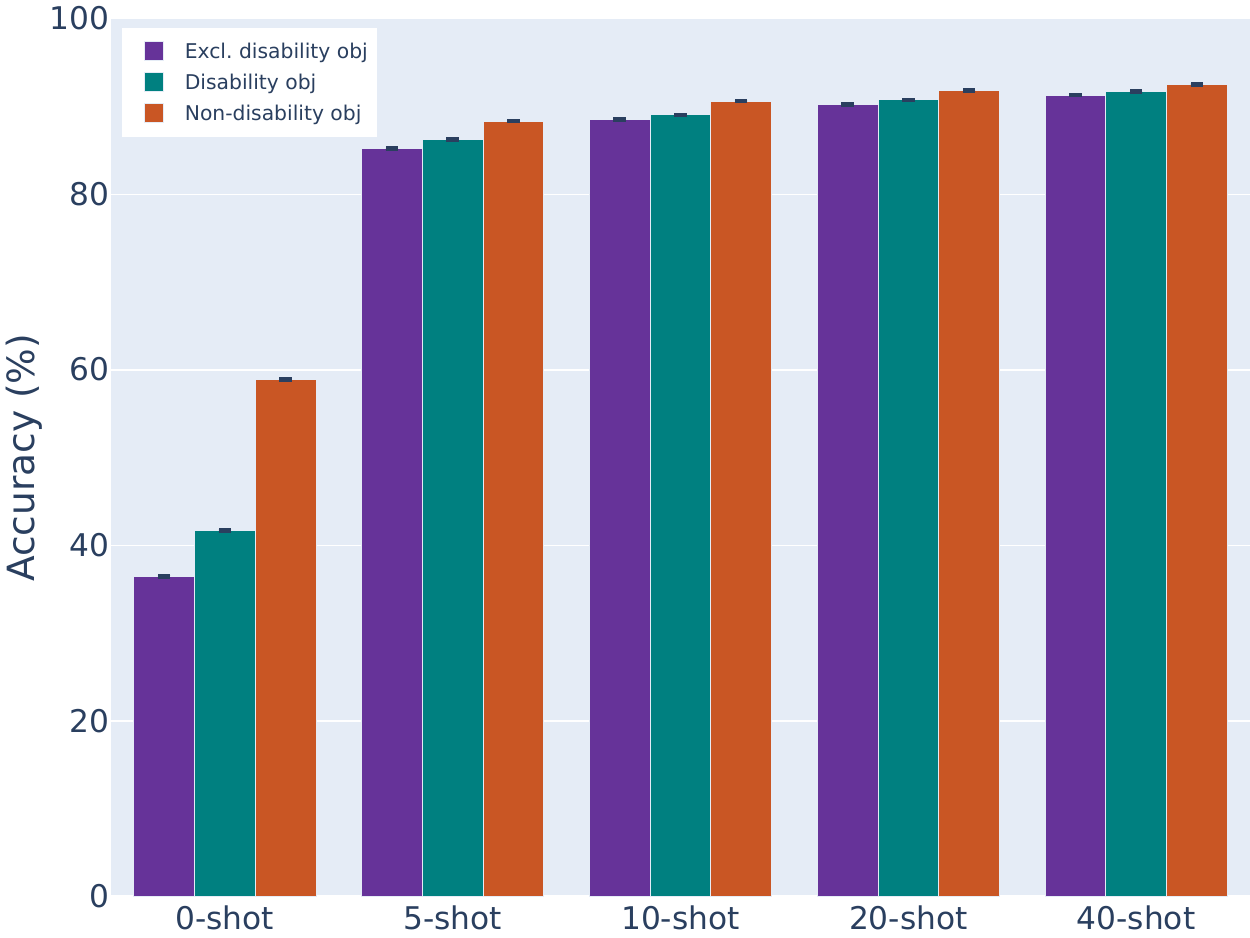}
        \caption{ORBIT Clean}
        \label{app:fig:orbit-clean-few-shot-by-num-shots}
    \end{subfigure}
    \hfill
    \begin{subfigure}{0.45\textwidth}
        \centering
        \includegraphics[width=\linewidth]{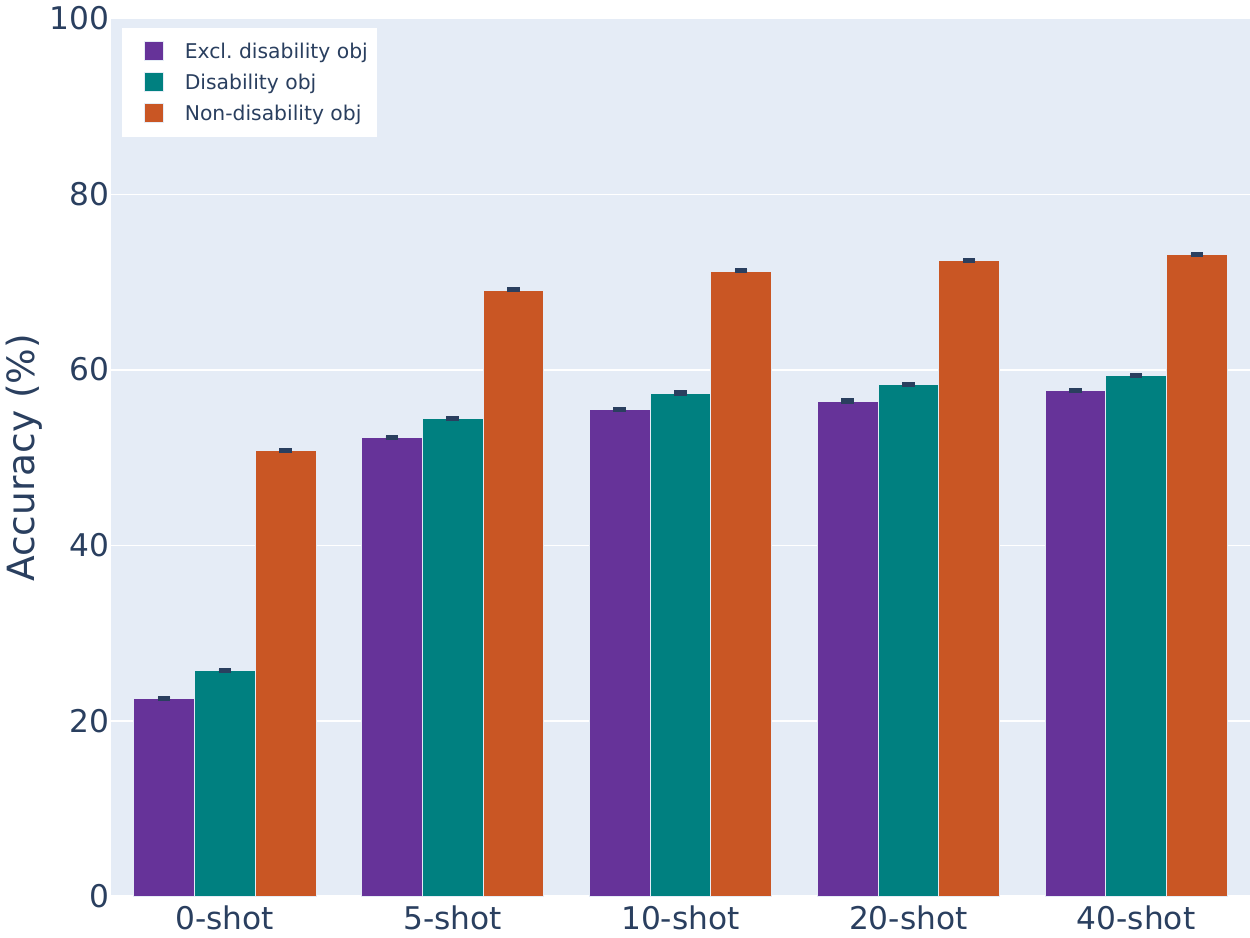}
        \caption{ORBIT Clutter}
        \label{app:fig:orbit-clutter-few-shot-by-num-shots}
    \end{subfigure}
    \caption{\textbf{A few-shot approach (ProtoNets~\cite{snell2017prototypical}) can reduce the accuracy gap between disability and non-disability objects, but not for realistic, cluttered images.} Bars represent the average accuracy (with 95\% c.i.) over all test frames for each shot setting ($K\!\!=\!\![5,10,20,40])$. $K\!\!=\!\!0$ is equivalent to the zero-shot setting described in~\cref{sec:image-content:dis-vs-non-dis}. Experimental details in~\cref{sec:image-content:few-shot}.}
    \label{app:fig:orbit-all-few-shot-by-num-shots}
\end{figure}

In~\cref{sec:image-content:few-shot} of the main paper, we show how a few-shot approach can be effective at reducing the accuracy difference between disability and non-disability objects in some scenarios.
We use ProtoNets~\cite{snell2017prototypical} as the few-shot approach, which computes an average embedding (or prototype) for each object class by simply averaging the embeddings of $K$ training images for each class.
A test image is then classified as the class whose prototype is most similar to the image’s embedding, where similarity is measured by Euclidean distance.
We extend~\cref{tab:orbit-clean-clutter-few-shot-dis-vs-non-dis-delta} in the main paper with~\cref{app:fig:orbit-clean-few-shot-by-num-shots,app:fig:orbit-clutter-few-shot-by-num-shots} here.
\cref{app:fig:orbit-clean-few-shot-by-num-shots} shows ProtoNets can reduce the accuracy difference between disability/exclusive disability and non-disability objects on the ORBIT Clean dataset.
\cref{app:fig:orbit-clutter-few-shot-by-num-shots} shows ProtoNets' results on the ORBIT Clutter dataset, however, the few-shot adaption is less effective, even with 40 shots per object.

\begin{figure}
    \centering
        \begin{subfigure}{0.45\textwidth}
        \includegraphics[width=\linewidth]{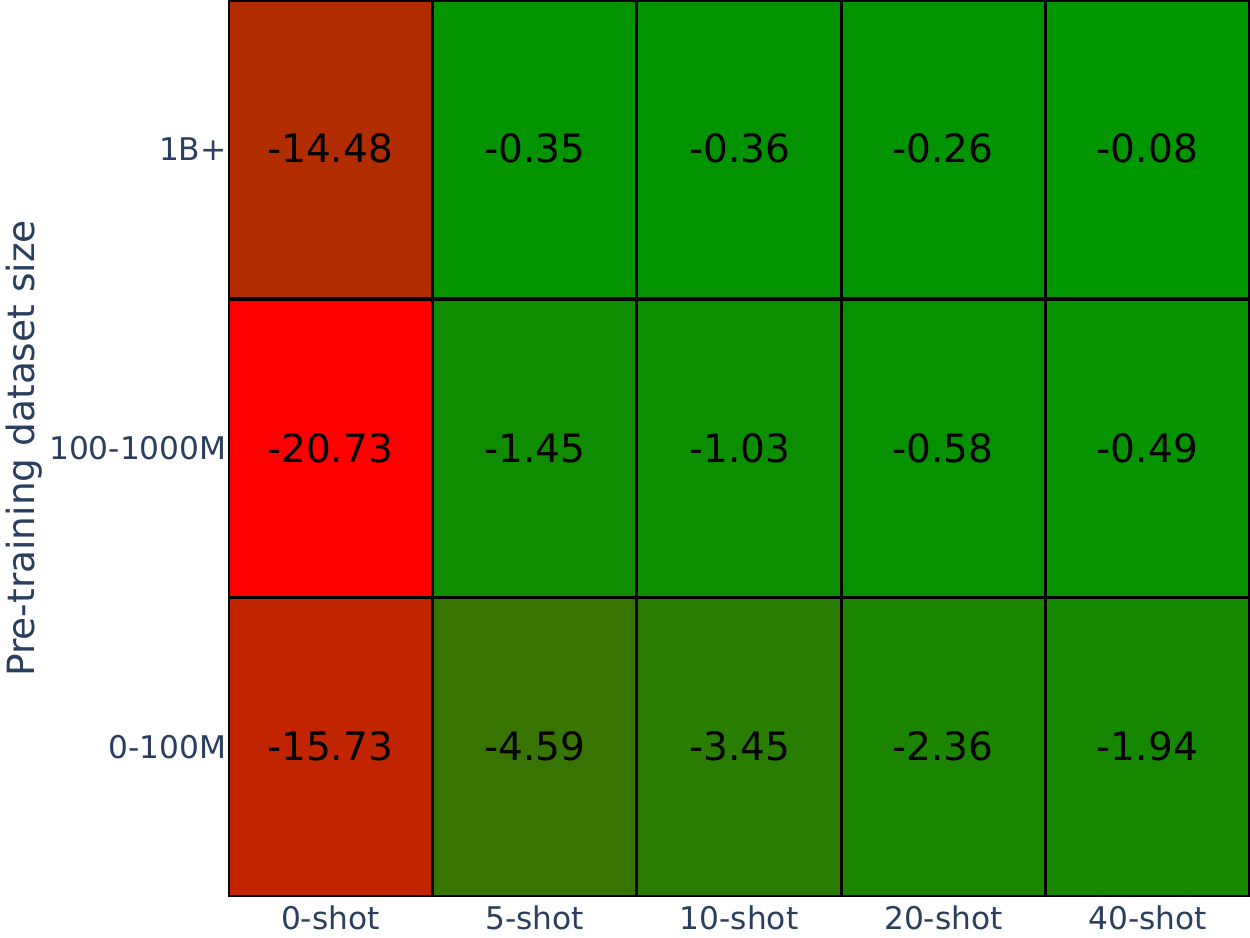}
        \caption{ORBIT Clean}
        \label{app:fig:orbit-clean-few-shot-by-dataset-size}
        \end{subfigure}
    \hfill
    \begin{subfigure}{0.45\textwidth}
        \centering
        \includegraphics[width=\linewidth]{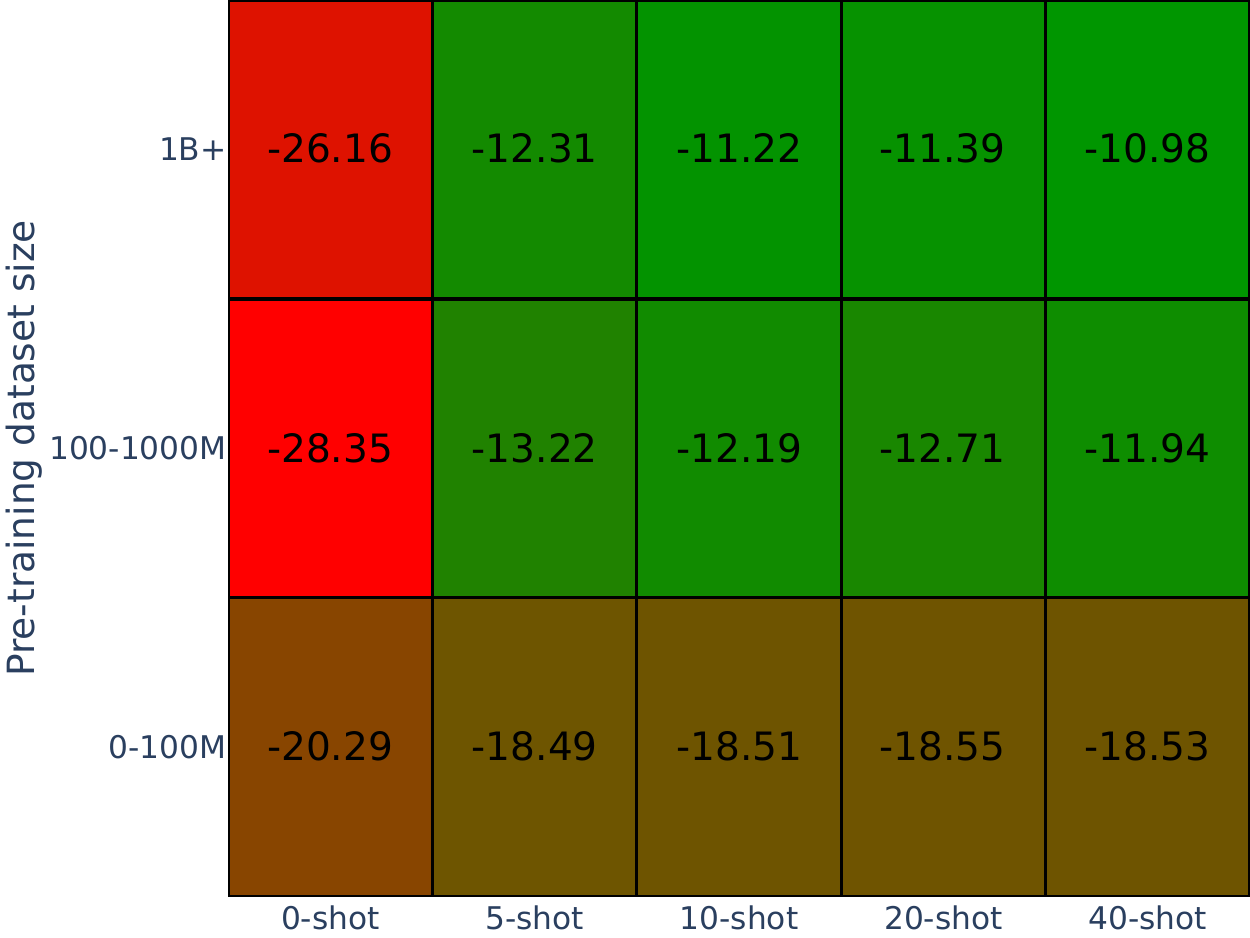}
        \caption{ORBIT Clutter}
        \label{app:fig:orbit-clutter-few-shot-by-dataset-size}
        \end{subfigure}
    \caption{\textbf{The larger the dataset used to pre-train CLIP, the more effective a few-shot approach is at closing the accuracy gap between disability and non-disability objects on ORBIT Clean, but this is less so for ORBIT Clutter especially for pre-training datasets \textless 100M examples}. Each block reports the average delta in accuracy between disability and non-disability objects for the models that fall within that group. Models: 25 CLIP variants.}
\end{figure}

In~\cref{app:fig:orbit-clean-few-shot-by-dataset-size,app:fig:orbit-clutter-few-shot-by-dataset-size}, we examine how CLIP's pre-training dataset size influences the few-shot adaptation on ORBIT Clean and Clutter, respectively.
We split the CLIP variants into three groups: those pre-trained on 0-100M examples, 100-1000M examples, and 1B+ examples.
For each group, we average the delta in accuracy between disability and non-disability objects for all CLIP variants in that group, for each shot setting.
%
%We see that in the zero-shot case, the average delta is largely consistent across the three groups, with the delta being 5-10 percentage larger for ORBIT Clutter than ORBIT Clean.
%
For ORBIT Clean, we see that as the pre-training dataset and the number of shots increase, the delta generally decreases -- with 1B+ pre-training examples and a 40-shot setting achieving the lowest delta  (-0.08) between disability and non-disability object accuracy.
For ORBIT Clutter, however, this trend is less pronounced.
Increasing the number of pre-training examples does reduce the delta generally, but the best setting (1B+ pre-training examples, 40 shots) still sees a delta of -10.98 percentage points.
Furthermore, for under 100M pre-training examples, the delta remains largely constant (around -18 percentage points) suggesting that a few-shot approach is less effective if the model has not seen enough pre-training data.

\subsection{Robustness to image quality from \gls{blv} users}

We include the raw marginal effects of each quality issue on model accuracy for all CLIP variants in~\cref{app:tab:orbit-vizwiz-reg-b16,app:tab:orbit-vizwiz-reg-b32,app:tab:orbit-vizwiz-reg-l14}.
These correspond to~\cref{fig:orbit-clean-vizwiz-classification-reg} in the main paper, with experimental details provided in~\cref{sec:image-quality-robustness}.
We note that the same image may be sampled multiple times as a result of the episodic sampling procedure (see~\cref{sec:prelims:zero-shot}).
No two tasks share the same set of $N$ objects, however, so for a given image, the model is always presented a different classification problem. 
The logistic regression is sensitive to input-output similarities, however, so we filter out all duplicate images to avoid biasing our sample.
This resulted in 93,698  images for ORBIT Clean and 6,764 for VizWiz-Classification.
We report the prevalence of each quality issue in these datasets in~\cref{app:tab:quality-issue-counts-orbit,app:tab:quality-issue-counts-vizwiz}.
\begin{table*}
\caption{Prevalence of each quality issue in the (a) ORBIT Clean and (b) VizWiz-Classification datasets. Numbers reported as the raw counts of each issue and as a percentage of the total non-disability/disability images (ORBIT Clean) and total images (VizWiz-Classification).}
\label{app:tab:quality-issue-counts}
\centering
\begin{subtable}[c]{\textwidth}
\scalebox{0.85}{
\begin{tabular}{l|cccccc}
\toprule
& \textbf{Total frames} & \textbf{Framing} & \textbf{Blur} & \textbf{Viewpoint} & \textbf{Occlusion} & \textbf{Lighting} \\
\midrule
Non-disability object & 86,185 & 52,275 (60.7\%) & 28,235 (32.8\%) &  14,253 (16.5\%) & 11,302 (13.1\%) & 3,788 (4.4\%)  \\
Disability object &  7,513 & 3,290 (43.8\%) & 2,237 (29.8\%) & 394 (5.2\%) & 976 (13.0\%) & 205 (2.7\%) \\
\bottomrule
\end{tabular}}
\caption{ORBIT Clean}
\label{app:tab:quality-issue-counts-orbit}
\end{subtable}
\qquad
\begin{subtable}[c]{\textwidth}
\scalebox{0.85}{
\begin{tabular}{l|C{1.5cm}C{1.5cm}C{1.5cm}C{1.5cm}C{1.5cm}C{1.55cm}C{1.55cm}C{1.3cm}}
\toprule
& \textbf{Total frames} & \textbf{Framing} & \textbf{Blur} & \textbf{Viewpoint} & \textbf{Occlusion} & \textbf{Overexposed} & \textbf{Underexposed} & \textbf{Other} \\ 
\midrule
Non-disability object & 6,764 & 3,715 (54.9\%) & 2,544 (37.6\%) & 1,118 (16.5\%) &  142 (2.1\%) & 327 (4.8\%) & 288 (4.3\%) &  16 (0.2\%) \\
\bottomrule
\end{tabular}}
\caption{VizWiz-Classification}
\label{app:tab:quality-issue-counts-vizwiz}
\end{subtable}
\end{table*}

\afterpage{%
\clearpage% Flush earlier floats (otherwise order might not be correct)
\begin{landscape}% Landscape page
\begin{table}
    \centering
    \caption{\textbf{Marginal effects of explanatory variables on CLIP's zero-shot classification accuracy (with ViT-B/16 vision encoders) on the ORBIT Clean and VizWiz-Classification datasets}. Main values are marginal effects, while values in brackets are p-values. */**/*** indicates within 90/95/99\% confidence interval, respectively. Experimental details in~\cref{sec:image-quality-robustness}. L-80M=LAION-80M, L-400M=LAION-400M, L-2B=LAION-2B, DC-L=DataComp-L, CP-L=CommonPool-L, CP-L-CLIP=CommonPool-L-CLIP.}
    \label{app:tab:orbit-vizwiz-reg-b16}
    \scalebox{0.85}{
\begin{tabular}{c|l|C{1.5cm}C{1.5cm}C{1.5cm}C{1.5cm}C{1.5cm}C{1.5cm}C{2.5cm}}
\toprule
\textbf{Dataset} & \textbf{Explanatory variable} & \textbf{ViT-B/16 (WIT)} & \textbf{ViT-B/16 (L-80M)} & \textbf{ViT-B/16 (L-400M)} & \textbf{ViT-B/16 (L-2B)} & \textbf{ViT-B/16 (DC-L)} & \textbf{ViT-B/16 (CP-L)} & \textbf{ViT-B/16 (CP-L-CLIP)} \\
\midrule
\multirow{22}{*}{\rotatebox[origin=c]{90}{\textbf{ORBIT Clean}}} & framing & 0.044*** & 0.044*** & 0.043*** & 0.042*** & 0.02*** & 0.054*** & 0.033*** \\
& & (1.0) & (1.0) & (1.0) & (1.0) & (1.0) & (1.0) & (1.0) \\
& blur & -0.095*** & -0.139*** & -0.117*** & -0.139*** & -0.153*** & -0.135*** & -0.146*** \\
& & (1.0) & (1.0) & (1.0) & (1.0) & (1.0) & (1.0) & (1.0) \\
& viewpoint & -0.115*** & -0.139*** & -0.122*** & -0.1*** & -0.092*** & -0.094*** & -0.086*** \\
& & (1.0) & (1.0) & (1.0) & (1.0) & (1.0) & (1.0) & (1.0) \\
& occlusion & -0.054*** & -0.099*** & -0.088*** & -0.086*** & -0.088*** & -0.104*** & -0.099*** \\
& & (1.0) & (1.0) & (1.0) & (1.0) & (1.0) & (1.0) & (1.0) \\
& lighting & -0.213*** & -0.212*** & -0.262*** & -0.245*** & -0.226*** & -0.297*** & -0.246*** \\
& & (1.0) & (1.0) & (1.0) & (1.0) & (1.0) & (1.0) & (1.0) \\
& excl. disability obj & -0.328*** & -0.341*** & -0.278*** & -0.245*** & -0.258*** & -0.352*** & -0.319*** \\
& & (1.0) & (1.0) & (1.0) & (1.0) & (1.0) & (1.0) & (1.0) \\
& excl. disability obj:framing & 0.143*** & 0.057*** & 0.108*** & 0.097*** & 0.141*** & 0.06*** & 0.097*** \\
& & (1.0) & (0.999) & (1.0) & (1.0) & (1.0) & (1.0) & (1.0) \\
& excl. disability obj:blur & 0.018 & 0.074*** & -0.015 & -0.017 & -0.021 & 0.044*** & 0.011 \\
& & (0.764) & (1.0) & (0.673) & (0.747) & (0.843) & (0.991) & (0.551) \\
& excl. disability obj:viewpoint & 0.113*** & 0.207*** & 0.173*** & 0.18*** & 0.069** & 0.07* & 0.065** \\
& & (1.0) & (1.0) & (1.0) & (1.0) & (0.969) & (0.947) & (0.959) \\
& excl. disability obj:occlusion & -0.058*** & 0.007 & -0.051** & -0.004 & -0.006 & 0.032 & 0.03 \\
& & (0.994) & (0.22) & (0.984) & (0.169) & (0.239) & (0.837) & (0.872) \\
& excl. disability obj:lighting & -0.239*** & -0.128** & -0.238*** & -0.208*** & -0.294*** & -0.204** & -0.214*** \\
& & (1.0) & (0.969) & (1.0) & (1.0) & (1.0) & (0.986) & (0.999) \\
 \midrule
\multirow{14}{*}{\rotatebox[origin=c]{90}{\textbf{VizWiz-Classification}}} & framing & 0.001 & 0.001 & -0.009 & -0.035*** & -0.01 & -0.028** & -0.018 \\
& & (0.051) & (0.093) & (0.542) & (0.995) & (0.599) & (0.979) & (0.862) \\
& blur & -0.028** & -0.019 & -0.009 & 0.0 & -0.014 & -0.015 & -0.02 \\
& & (0.976) & (0.867) & (0.52) & (0.006) & (0.727) & (0.761) & (0.885) \\
& rotation & -0.079*** & -0.116*** & -0.055*** & -0.086*** & -0.07*** & -0.092*** & -0.09*** \\
& & (1.0) & (1.0) & (0.999) & (1.0) & (1.0) & (1.0) & (1.0) \\
& occlusion & -0.096** & -0.144*** & -0.138*** & -0.187*** & -0.142*** & -0.209*** & -0.186*** \\
& & (0.978) & (0.999) & (0.999) & (1.0) & (0.999) & (1.0) & (1.0) \\
& overexposure & -0.034 & 0.023 & -0.033 & -0.011 & -0.07** & -0.064** & -0.029 \\
& & (0.777) & (0.592) & (0.758) & (0.301) & (0.987) & (0.974) & (0.688) \\
& underexposure & -0.02 & -0.088*** & -0.084*** & -0.065** & -0.051* & -0.073** & -0.074** \\
& & (0.498) & (0.996) & (0.995) & (0.969) & (0.91) & (0.984) & (0.985) \\
& other & 0.129 & 0.099 & 0.009 & 0.243* & 0.017 & 0.115 & 0.04 \\
& & (0.668) & (0.579) & (0.059) & (0.911) & (0.112) & (0.63) & (0.251) \\
 \bottomrule
\end{tabular}}
\end{table}

\clearpage

\begin{table}
    \centering
    \caption{\textbf{Marginal effects of explanatory variables on CLIP's zero-shot classification accuracy (with ViT-B/32 vision encoders) on the ORBIT Clean and VizWiz-Classification datasets}. Main values are marginal effects, while values in brackets are p-values. */**/*** indicates within 90/95/99\% confidence interval, respectively. Experimental details in~\cref{sec:image-quality-robustness}. L-80M=LAION-80M, L-400M=LAION-400M, L-2B=LAION-2B, DC-S=DataComp-S, DC-M=DataComp-M, CP-S=CommonPool-S, CP-S-CLIP=CommonPool-S-CLIP, CP-M=CommonPool-M, CP-M-CLIP=CommonPool-M-CLIP.}
    \label{app:tab:orbit-vizwiz-reg-b32}
    \scalebox{0.85}{
\begin{tabular}{c|l|C{1.5cm}C{1.5cm}C{1.5cm}C{1.5cm}C{1.5cm}C{1.5cm}C{1.5cm}C{2.5cm}C{1.5cm}C{2.5cm}}
\toprule
\textbf{Dataset} & \textbf{Explanatory variable} & \textbf{ViT-B/32 (WIT)} & \textbf{ViT-B/32 (L-80M)} & \textbf{ViT-B/32 (L-400M)} & \textbf{ViT-B/32 (L-2B)} & \textbf{ViT-B/32 (DC-S)} & \textbf{ViT-B/32 (DC-M)} & \textbf{ViT-B/32 (CP-S)} & \textbf{ViT-B/32 (CP-S-CLIP)} & \textbf{ViT-B/32 (CP-M)} & \textbf{ViT-B/32 (CP-M-CLIP)} \\
\hline
\multirow{22}{*}{\rotatebox[origin=c]{90}{\textbf{ORBIT Clean}}} & framing & 0.062*** & 0.046*** & 0.054*** & 0.051*** & 0.035*** & 0.101*** & 0.046*** & 0.05*** & 0.11*** & 0.097*** \\
& & (1.0) & (1.0) & (1.0) & (1.0) & (1.0) & (1.0) & (1.0) & (1.0) & (1.0) & (1.0) \\
& blur & -0.106*** & -0.128*** & -0.142*** & -0.132*** & -0.025*** & -0.113*** & -0.029*** & -0.05*** & -0.103*** & -0.119*** \\
&  & (1.0) & (1.0) & (1.0) & (1.0) & (1.0) & (1.0) & (1.0) & (1.0) & (1.0) & (1.0) \\
& viewpoint & -0.096*** & -0.111*** & -0.111*** & -0.099*** & -0.036*** & -0.058*** & -0.011*** & -0.032*** & -0.045*** & -0.092*** \\
& & (1.0) & (1.0) & (1.0) & (1.0) & (1.0) & (1.0) & (0.999) & (1.0) & (1.0) & (1.0) \\
& occlusion & -0.075*** & -0.095*** & -0.094*** & -0.088*** & -0.088*** & -0.142*** & -0.072*** & -0.08*** & -0.109*** & -0.123*** \\
& & (1.0) & (1.0) & (1.0) & (1.0) & (1.0) & (1.0) & (1.0) & (1.0) & (1.0) & (1.0) \\
& lighting & -0.174*** & -0.297*** & -0.292*** & -0.261*** & -0.117*** & -0.228*** & -0.209*** & -0.217*** & -0.296*** & -0.294*** \\
& & (1.0) & (1.0) & (1.0) & (1.0) & (1.0) & (1.0) & (1.0) & (1.0) & (1.0) & (1.0) \\
& excl. disability obj & -0.33*** & -0.359*** & -0.311*** & -0.265*** & -0.297*** & -0.311*** & -0.124*** & -0.204*** & -0.415*** & -0.576*** \\
& & (1.0) & (1.0) & (1.0) & (1.0) & (1.0) & (1.0) & (1.0) & (1.0) & (1.0) & (1.0) \\
& excl. disability obj:framing & 0.137*** & 0.118*** & 0.124*** & 0.113*** & -0.016 & -0.094*** & -0.259*** & -0.147*** & -0.064*** & 0.049* \\
& & (1.0) & (1.0) & (1.0) & (1.0) & (0.479) & (1.0) & (1.0) & (1.0) & (0.99) & (0.949) \\
& excl. disability obj:blur & 0.072*** & -0.012 & 0.011 & 0.013 & 0.028 & 0.045** & -0.01 & 0.043* & 0.062** & 0.062** \\
& & (1.0) & (0.458) & (0.479) & (0.593) & (0.759) & (0.959) & (0.347) & (0.948) & (0.987) & (0.984) \\
& excl. disability obj:viewpoint & 0.128*** & 0.166*** & 0.06* & 0.129*** & 0.219*** & 0.143*** & 0.148** & 0.225*** & 0.259*** & 0.268*** \\
& & (1.0) & (1.0) & (0.911) & (1.0) & (1.0) & (0.999) & (0.99) & (1.0) & (1.0) & (1.0) \\
& excl. disability obj:occlusion & -0.151*** & 0.109*** & -0.022 & -0.025 & 0.176*** & 0.036 & -0.013 & 0.038 & 0.166*** & 0.31*** \\
& & (1.0) & (1.0) & (0.659) & (0.779) & (1.0) & (0.758) & (0.313) & (0.791) & (1.0) & (1.0) \\
& excl. disability obj:lighting & -0.188*** & 0.025 & -0.255*** & -0.285*** & 0.17*** & -0.313** & 0.188*** & 0.123* & -0.156 & -0.273* \\
& & (1.0) & (0.312) & (0.998) & (1.0) & (1.0) & (0.98) & (0.998) & (0.945) & (0.768) & (0.904) \\
\midrule
\multirow{14}{*}{\rotatebox[origin=c]{90}{\textbf{VizWiz-Classification}}} & framing & 0.011 & -0.024** & -0.002 & -0.034*** & -0.015* & -0.034*** & 0.014* & -0.005 & -0.021* & -0.003 \\
& & (0.634) & (0.963) & (0.115) & (0.995) & (0.945) & (0.997) & (0.908) & (0.441) & (0.936) & (0.19) \\
& blur & -0.009 & -0.003 & -0.029** & 0.007 & -0.005 & -0.026** & -0.001 & -0.02** & -0.033*** & -0.03** \\
& & (0.517) & (0.202) & (0.98) & (0.433) & (0.455) & (0.97) & (0.106) & (0.97) & (0.995) & (0.984) \\
& rotation & -0.072*** & -0.153*** & -0.052*** & -0.113*** & -0.067*** & -0.092*** & -0.075*** & -0.117*** & -0.16*** & -0.111*** \\
& & (1.0) & (1.0) & (0.998) & (1.0) & (1.0) & (1.0) & (1.0) & (1.0) & (1.0) & (1.0) \\
& occlusion & -0.144*** & -0.121*** & -0.17*** & -0.132*** & -0.057* & -0.175*** & -0.098** & -0.128*** & -0.182*** & -0.216*** \\
& & (0.999) & (0.993) & (1.0) & (0.997) & (0.907) & (1.0) & (0.987) & (0.996) & (1.0) & (1.0) \\
& overexposure & 0.035 & -0.063** & -0.059** & -0.025 & -0.004 & -0.082*** & -0.013 & -0.027 & -0.11*** & -0.063** \\
&  & (0.775) & (0.974) & (0.961) & (0.618) & (0.17) & (0.995) & (0.489) & (0.778) & (1.0) & (0.972) \\
& underexposure & -0.009 & -0.034 & -0.095*** & -0.046 & -0.019 & -0.081*** & -0.008 & -0.049* & -0.011 & -0.094*** \\
& & (0.237) & (0.754) & (0.998) & (0.868) & (0.648) & (0.992) & (0.307) & (0.948) & (0.294) & (0.997) \\
& other & -0.13 & 0.176 & 0.06 & 0.063 & 0.002 & 0.084 & -0.106 & 0.107 & -0.101 & 0.016 \\
& & (0.705) & (0.877) & (0.368) & (0.388) & (0.023) & (0.543) & (0.635) & (0.876) & (0.577) & (0.103) \\
 \bottomrule
\end{tabular}}
\end{table}

\clearpage

\begin{table}
    \centering
    \caption{\textbf{Marginal effects of explanatory variables on CLIP's zero-shot classification accuracy (with ViT-L/14, ViT-H/14 and ViT-g/14 vision encoders) on the ORBIT Clean and VizWiz-Classification datasets}. Main values are marginal effects, while values in brackets are p-values. */**/*** indicates within 90/95/99\% confidence interval, respectively. Experimental details in~\cref{sec:image-quality-robustness}. L-80M=LAION-80M, L-400M=LAION-400M, L-2B=LAION-2B, DC-Xl=DataComp-XL, CP-XL-CLIP=CommonPool-XL-CLIP.}
    \label{app:tab:orbit-vizwiz-reg-l14}
    \scalebox{0.85}{
\begin{tabular}{c|l|C{1.5cm}C{1.5cm}C{1.5cm}C{1.5cm}C{1.5cm}C{2.5cm}C{1.5cm}C{1.5cm}}
\toprule
\textbf{Dataset} & \textbf{Explanatory variable} & \textbf{ViT-L/14 (WIT)} & \textbf{ViT-L/14 (L-80M)} & \textbf{ViT-L/14 (L-400M)} & \textbf{ViT-L/14 (L-2B)} & \textbf{ViT-L/14 (DC-XL)} & \textbf{ViT-L/14 (CP-XL-CLIP)} & \textbf{ViT-H/14 (L-2B)} & \textbf{ViT-g/14 (L-2B)} \\
\hline
\multirow{22}{*}{\rotatebox[origin=c]{90}{\textbf{ORBIT Clean}}} & framing & -0.013*** & 0.027*** & 0.03*** & 0.019*** & -0.016*** & -0.018*** & 0.025*** & 0.025*** \\
&  & (1.0) & (1.0) & (1.0) & (1.0) & (1.0) & (1.0) & (1.0) & (1.0) \\
& blur & -0.098*** & -0.139*** & -0.131*** & -0.129*** & -0.099*** & -0.114*** & -0.113*** & -0.123*** \\
& & (1.0) & (1.0) & (1.0) & (1.0) & (1.0) & (1.0) & (1.0) & (1.0) \\
& viewpoint & -0.117*** & -0.15*** & -0.114*** & -0.124*** & -0.087*** & -0.09*** & -0.115*** & -0.106*** \\
& & (1.0) & (1.0) & (1.0) & (1.0) & (1.0) & (1.0) & (1.0) & (1.0) \\
& occlusion & -0.065*** & -0.091*** & -0.085*** & -0.078*** & -0.068*** & -0.068*** & -0.078*** & -0.082*** \\
& & (1.0) & (1.0) & (1.0) & (1.0) & (1.0) & (1.0) & (1.0) & (1.0) \\
& lighting & -0.174*** & -0.288*** & -0.248*** & -0.223*** & -0.189*** & -0.184*** & -0.173*** & -0.222*** \\
& & (1.0) & (1.0) & (1.0) & (1.0) & (1.0) & (1.0) & (1.0) & (1.0) \\
& excl. disability obj & -0.26*** & -0.235*** & -0.257*** & -0.258*** & -0.223*** & -0.245*** & -0.267*** & -0.283*** \\
& & (1.0) & (1.0) & (1.0) & (1.0) & (1.0) & (1.0) & (1.0) & (1.0) \\
& excl. disability obj:framing & 0.085*** & 0.138*** & 0.085*** & 0.112*** & 0.16*** & 0.127*** & 0.103*** & 0.138*** \\
& & (1.0) & (1.0) & (1.0) & (1.0) & (1.0) & (1.0) & (1.0) & (1.0) \\
& excl. disability obj:blur & -0.012 & -0.04** & -0.013 & -0.032** & -0.083*** & -0.051*** & -0.039*** & -0.027** \\
& & (0.644) & (0.982) & (0.635) & (0.98) & (1.0) & (1.0) & (0.997) & (0.956) \\
& excl. disability obj:viewpoint & 0.101*** & 0.113*** & 0.144*** & 0.108*** & 0.041 & 0.119*** & 0.125*** & 0.118*** \\
& & (1.0) & (0.999) & (1.0) & (1.0) & (0.884) & (1.0) & (1.0) & (1.0) \\
& excl. disability obj:occlusion & -0.052*** & -0.023 & 0.014 & -0.005 & -0.004 & -0.018 & 0.011 & 0.041** \\
& & (0.998) & (0.696) & (0.56) & (0.237) & (0.218) & (0.765) & (0.492) & (0.982) \\
& excl. disability obj:lighting & -0.192*** & -0.369*** & -0.185*** & -0.147*** & -0.133*** & -0.153*** & -0.176*** & -0.243*** \\
& & (1.0) & (1.0) & (0.999) & (0.998) & (1.0) & (1.0) & (1.0) & (1.0) \\
 \midrule
\multirow{14}{*}{\rotatebox[origin=c]{90}{\textbf{VizWiz-Classification}}} & framing & 0.007 & -0.04*** & -0.038*** & -0.014 & -0.002 & 0.013 & -0.006 & -0.008 \\
& & (0.434) & (0.999) & (0.998) & (0.748) & (0.138) & (0.737) & (0.402) & (0.503) \\
& blur & -0.005 & 0.015 & -0.014 & -0.002 & -0.007 & -0.009 & -0.012 & -0.019 \\
& & (0.317) & (0.773) & (0.737) & (0.131) & (0.404) & (0.552) & (0.676) & (0.87) \\
& rotation & -0.052*** & -0.093*** & -0.076*** & -0.031* & -0.065*** & -0.048*** & -0.066*** & -0.101*** \\
& & (0.999) & (1.0) & (1.0) & (0.948) & (1.0) & (0.998) & (1.0) & (1.0) \\
& occlusion & -0.118*** & -0.074* & -0.143*** & -0.126*** & -0.134*** & -0.126*** & -0.136*** & -0.106** \\
& & (0.996) & (0.915) & (0.999) & (0.998) & (0.999) & (0.999) & (0.999) & (0.987) \\
& overexposure & -0.004 & -0.017 & -0.052* & 0.001 & 0.015 & -0.018 & 0.019 & 0.004 \\
& & (0.105) & (0.449) & (0.934) & (0.015) & (0.416) & (0.499) & (0.499) & (0.101) \\
& underexposure & -0.023 & -0.034 & -0.092*** & -0.076*** & -0.023 & -0.056* & -0.035 & -0.028 \\
& & (0.57) & (0.743) & (0.998) & (0.99) & (0.562) & (0.949) & (0.761) & (0.642) \\
& other & 0.093 & 0.281** & 0.08 & 0.194 & 0.235 & 0.051 & 0.066 & 0.022 \\
& & (0.525) & (0.972) & (0.473) & (0.83) & (0.876) & (0.314) & (0.395) & (0.137) \\ 
 \bottomrule
\end{tabular}}
\end{table}

\end{landscape}
\clearpage% Flush page
}

\subsection{Robustness to language used by \gls{blv} users}
\label{app:sec:textual-content}

\subsubsection{CLIP classifies objects more accurately when they are described by color rather than material}
\label{app:sec:textual-content:color-vs-material}

\vspace{-0.1em}
We extend~\cref{tab:orbit-clean-avg-clip-score-per-prompt-type} in the main paper, with~\cref{app:tab:orbit-clutter-avg-clip-score-per-prompt-type} for the ORBIT Clutter dataset here.
We see that the CLIP scores for the lower bound prompt (\ie just the object name) are the highest, followed by the color prompt.
Similar to~\cref{tab:orbit-clean-avg-clip-score-per-prompt-type}, we see that both the material prompt and the upper bound prompt which includes the object's material have the lowest CLIP scores, suggesting that including the object's material in the prompt harms embedding alignment.
\begin{table}
\caption{\textbf{Including an object's material in its prompt leads to text embeddings that are the least aligned with the object's image embeddings.} CLIP scores~\cite{hessel2021clipscore} between image and prompt embeddings are averaged (with 95\% c.i.) for 100 images per object per prompt type on ORBIT Clutter.}
\centering
\vspace{-1em}
    \scalebox{0.8}{
    \begin{tabular}{l|C{1.2cm}|C{1.6cm}|C{1.6cm}|C{1.7cm}}
    \toprule
         \textbf{Prompt} & \textbf{Obj. name} & \textbf{Material + obj. name} & \textbf{Color + obj. name} & \textbf{Color + material + obj. name}  \\ \hline
         \textbf{CLIP Score} & \textbf{22.81} $\pm$ \textbf{0.02} & 21.92 $\pm$ 0.02 & 22.73 $\pm$ 0.02 & 21.86 $\pm$ 0.02 \\
    \bottomrule
    \end{tabular}}
    \label{app:tab:orbit-clutter-avg-clip-score-per-prompt-type}
    \vspace{-1em}
\end{table}

\begin{figure}
    \centering
    \includegraphics[width=0.95\linewidth]{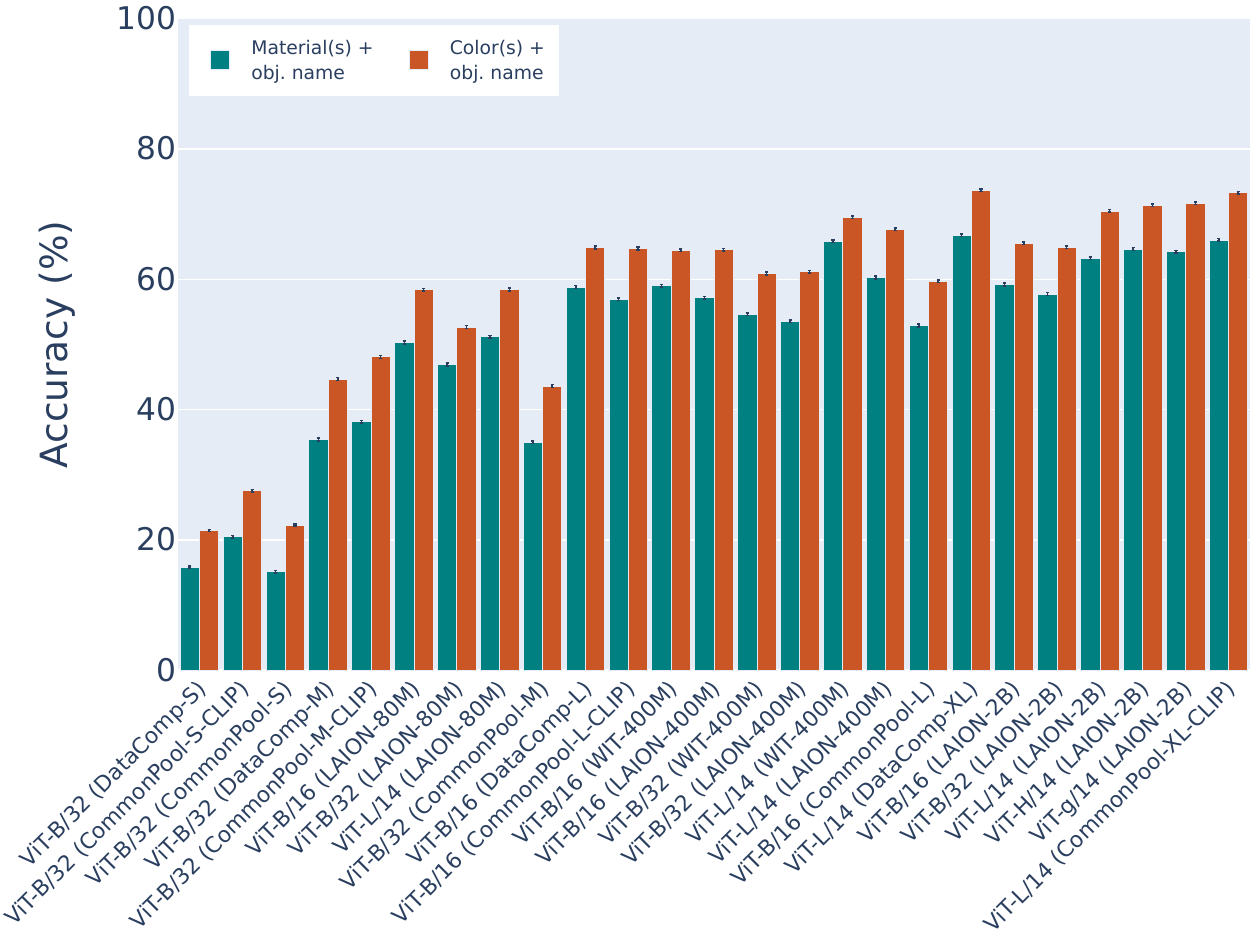}
    \caption{\textbf{All CLIP variants classify objects more accurately when objects are described by their color rather than their material}. Each bar is the average accuracy (with 95\% c.i.) over 200K images (100K ORBIT Clean, 100K ORBIT Clutter) for that CLIP variant when given either a material or color prompt. Variants ordered by pre-training dataset size.}
    \label{app:fig:orbit-clean-orbit-clutter-avg-acc-per-prompt-type-by-model}
    \vspace{-1em}
\end{figure}

\begin{figure}
    \centering
    \includegraphics[width=0.9\linewidth]{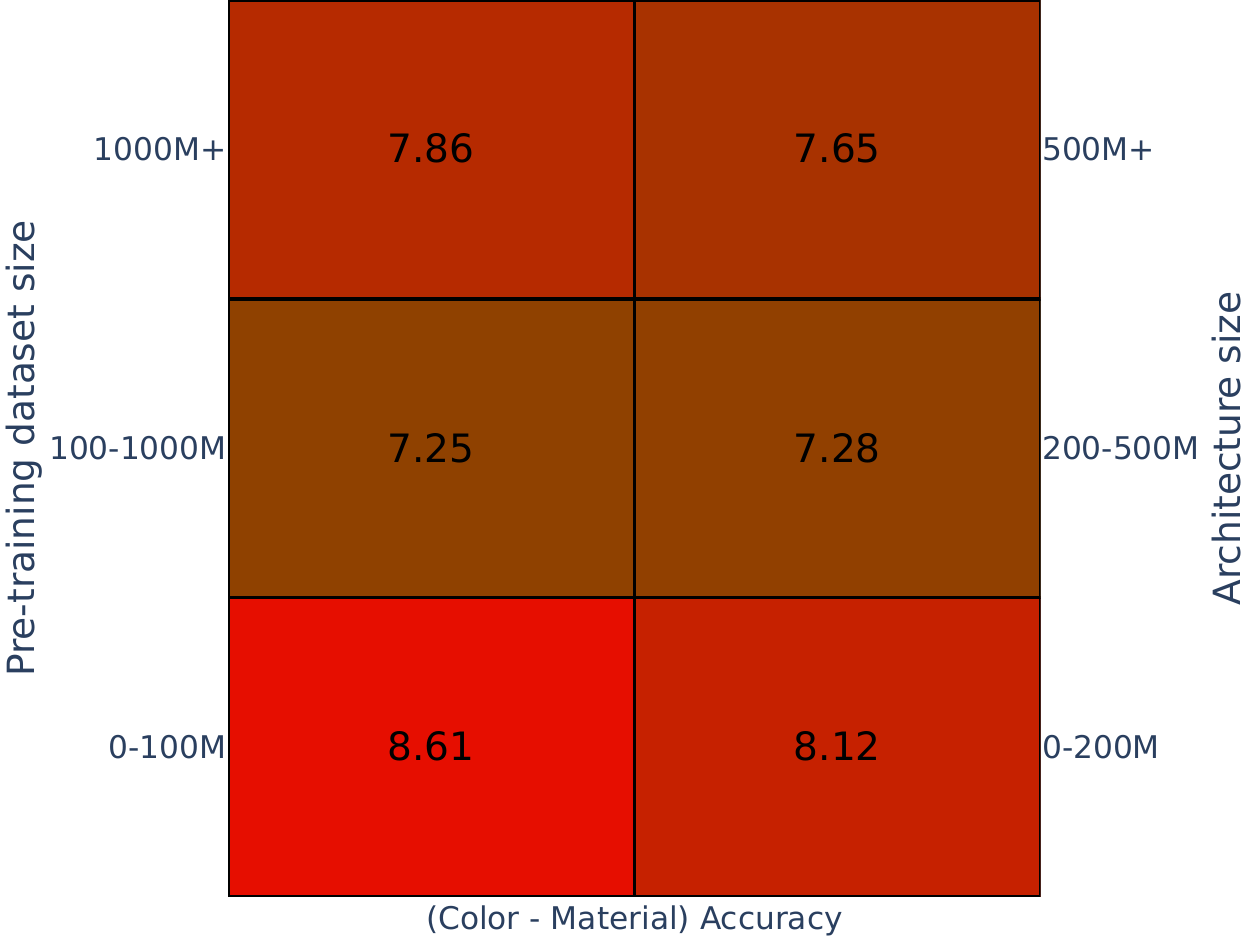}
    \caption{\textbf{Increasing the pre-training dataset and architecture size only marginally reduces the difference in zero-shot accuracy between prompts describing an object by its color versus material.} Numbers reported are the delta in zero-shot accuracy between when a color versus material prompt is used as the text input to CLIP on the ORBIT Clean dataset. Each block averaged the delta for all CLIP variants that fall within that group. Experimental details in~\cref{sec:text-content:color-vs-material}. Models: All 25 CLIP variants}
    \label{app:fig:orbit-clean-delta-color-material-acc-by-model-dataset-size-grid}
    \vspace{-1em}
\end{figure}

We explore the impact this has on classifier accuracy by combining the textual prompts with the standard zero-shot set-up described in~\cref{sec:prelims:zero-shot}.
Specifically, rather than embedding the raw ORBIT object labels for each task's $N$ classes, we instead embed their textual prompts.
In the first experiment, we embed all $N$ objects as their color prompts, and in the second as their material prompts.
For both experiments, we use $T=50$, $N=20$, $M=100$.
In~\cref{app:fig:orbit-clean-orbit-clutter-avg-acc-per-prompt-type-by-model}, we see that across all CLIP variants, objects are classified more accurately when they are described by their color rather than their material -- by 7.1 percentage points more, on average.
We see that this difference is largely constant regardless of both architecture and pre-training dataset size (see~\cref{app:fig:orbit-clean-delta-color-material-acc-by-model-dataset-size-grid}).

\vspace{-1em}
\section{Example-based analysis}
\label{app:sec:example-analysis}

\subsection{Standardized image selection}

We run our analysis on 180 images spanning 20 objects which are selected through a standardized process as a way to systematically assess failure cases.
Specifically, we select the 5 top- and bottom-performing (disability and non-disability) objects from the ORBIT dataset using the standardized zero-shot classification set-up (see objects in~\cref{app:tab:image-analysis:objects}).
We take performance to be the average accuracy per object, computed over all the CLIP variants we considered.
For each object, we extract the noun phrase from its raw label and apply simple pre-processing to ensure that it is unambiguous and concise (see cleaned phrases in~\cref{app:tab:image-analysis:objects}).
%
%In cases where this noun phrase was ambiguous (\eg grinder, white cane), an adjective was added to better specify the object based on the images available of that object (\eg tobacco grinder, guide cane).
%
%Any typos and extraneous adjectives were also removed (\eg folded long guide cane -- guide cane).
These cleaned noun phrases are used as the text prompts for all three downstream models we study.
\begin{table}[]
    \centering
    \caption{\textbf{Top- and bottom-performing disability and non-disability objects from the ORBIT dataset sed for the example-based analysis.} The noun phrases are extracted from the raw label and cleaned to ensure they are unambiguous and concise.}
    \label{app:tab:image-analysis:objects}
    \vspace{-0.5em}
    \scalebox{0.9}{
    \begin{tabular}{cc|l|l}
    \toprule
        & & \textbf{Raw object label} & \textbf{Cleaned noun phrase} \\
        \midrule
         \multirow{10}{*}{\rotatebox[origin=c]{90}{\textbf{Disability objects}}} & \multirow{5}{*}{\rotatebox[origin=c]{90}{Top 5}} & my braille displat & braille sense display \\
        & & dog poo & dog poo\\
        & & dog lead & dog lead\\
        & & white cane & guide cane\\
        & & folded long guide cane & guide cane\\
        & \multirow{5}{*}{\rotatebox[origin=c]{90}{Bottom 5}} & victor reader stream & victor reader stream\\
        & & braille note & braille notetaker\\
        & & liquid level indicator & liquid level indicator\\
        & & liquid level indicator & liquid level indicator\\
        & & dictaphone & dictaphone\\
        \hline
    \multirow{10}{*}{\rotatebox[origin=c]{90}{\textbf{Non-disability objects}}} & \multirow{5}{*}{\rotatebox[origin=c]{90}{Top 5}} & back patio gate & gate\\
    & & local post box & post box\\
    & & wine glass & wine glass\\
    & & tv remote control	& remote control\\
    & & remote control & remote control\\
    & \multirow{5}{*}{\rotatebox[origin=c]{90}{Bottom 5}} & digital dab radio	& digital radio\\
    & & my clock & digital clock\\
    & & grinder & tobacco grinder\\
    & & dog streetball & ball\\
    & & shoulder bag & shoulder bag\\
\bottomrule
    \end{tabular}}
    \vspace{-0.7em}
\end{table}
We then sample 9 images for each of the 20 objects -- 6 from its clutter videos and 3 from its clean videos. We only sample images where the object is tagged as present.
To increase image diversity, we ensure that images are sampled from all videos available for each object, and are sampled at even intervals. 
Specifically, for clean videos we sample the 3 frames at 25\%, 50\% and 75\% positions, alternating the video we sample from each time (\eg if an object has 2 clean videos, then we sample 1 frame at 25\% of video 1, 1 frame at 50\% of video 2, and 1 frame at 75\% of video 1).
For clutter videos, we sample 6 frames at 25\%, 35\%, 45\%, 55\%, 65\% and 75\%, also alternating the video for each sample.
We limit frame sampling to between 25\% and 75\% of each video as ORBIT data collectors were instructed to start each video with the camera close to the object and then move it further away, so we wanted to exclude frames where the camera might be too close/far from the object.

\subsection{Object detection with OWL-ViT}

\begin{figure*}
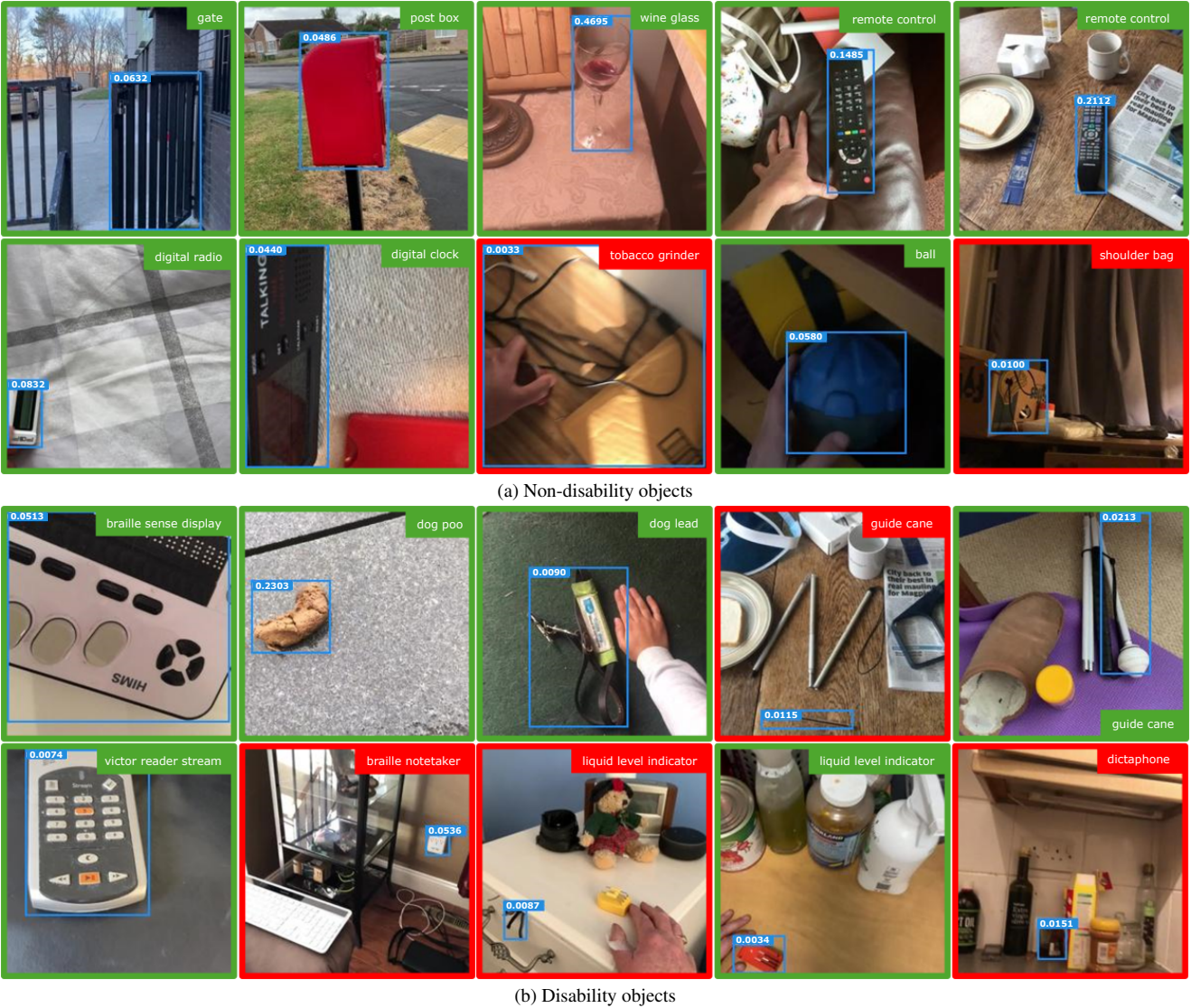

    \centering
    \begin{subfigure}[b]{\textwidth}
        \includesvg[width=\textwidth,inkscapelatex=false]{plots/image-analysis-det2a.svg}
        \caption{Non-disability objects}
        \label{app:fig:image-analysis:det2a}
    \end{subfigure}
    \hfill
    \begin{subfigure}[b]{\textwidth}
        \includesvg[width=\textwidth,inkscapelatex=false]{plots/image-analysis-det2b.svg}
        \caption{Disability objects}
        \label{app:fig:image-analysis:det2b}
    \end{subfigure}
    \caption{\textbf{OWL-ViT detects disability objects less confidently than non-disability objects.} For each of the (a) 10 non-disability and (b) 10 disability objects, we show the image with the highest-scoring bounding box out of the 9 images analyzed for that object.}
    \label{app:fig:image-analysis:det2}
    \vspace{-1em}
\end{figure*}

We extend~\cref{fig:image-analysis:det1} in the main paper with~\cref{app:fig:image-analysis:det2} here, where we show one example of OWL-ViT's bounding box detections for each of the 10 disability and 10 non-disability objects.
Specifically, for each object we show the image that had the bounding box with the highest confidence score across all 9 images analyzed for that object.
We see that the confidence scores in these images are $\sim$3x lower for disability than non-disability objects, on average.
We also see that for 4/10 disability objects, the incorrect object is detected (versus 2/10 non-disability objects).

We also report the mean intersection-over-union (IOU) between OWL-ViT's predicted and ground-truth bounding box for each object in~\cref{app:tab:image-analysis-det3}.
Since ground-truth bounding boxes are only publicly available for the clutter images, we manually annotated the remaining 6 clean images per object.
Our results show that the mean IOU is $\sim$2x lower for disability compared to non-disability objects.
Taken together, these results suggest that overall, OWL-ViT performs less reliably and confidently for disability content.
\begin{table}[]
    \centering
    \caption{\textbf{OWL-ViT's mean intersection-over-union (IOU) is $\sim$2x lower for disability compared to non-disability objects.} Mean IOU (with 95\% c.i.) is computed between the predicted and ground-truth bounding box for each object.}
    \vspace{-0.5em}
    \scalebox{0.9}{
    \begin{tabular}{l|c}
        \toprule
        & \textbf{mean IOU} \\
        \midrule
         Disability objects & 0.1323 (0.0947) \\
         Non-disability objects & \textbf{0.2488 (0.1829)} \\
         \bottomrule
    \end{tabular}}
    \label{app:tab:image-analysis-det3}
    \vspace{-1.2em}
\end{table}

\subsection{Semantic segmentation with CLIPSeg}
\label{app:sec:example-analysis:seg}

\begin{figure*}
    \centering
    \includesvg[width=0.8\linewidth,inkscapelatex=false]{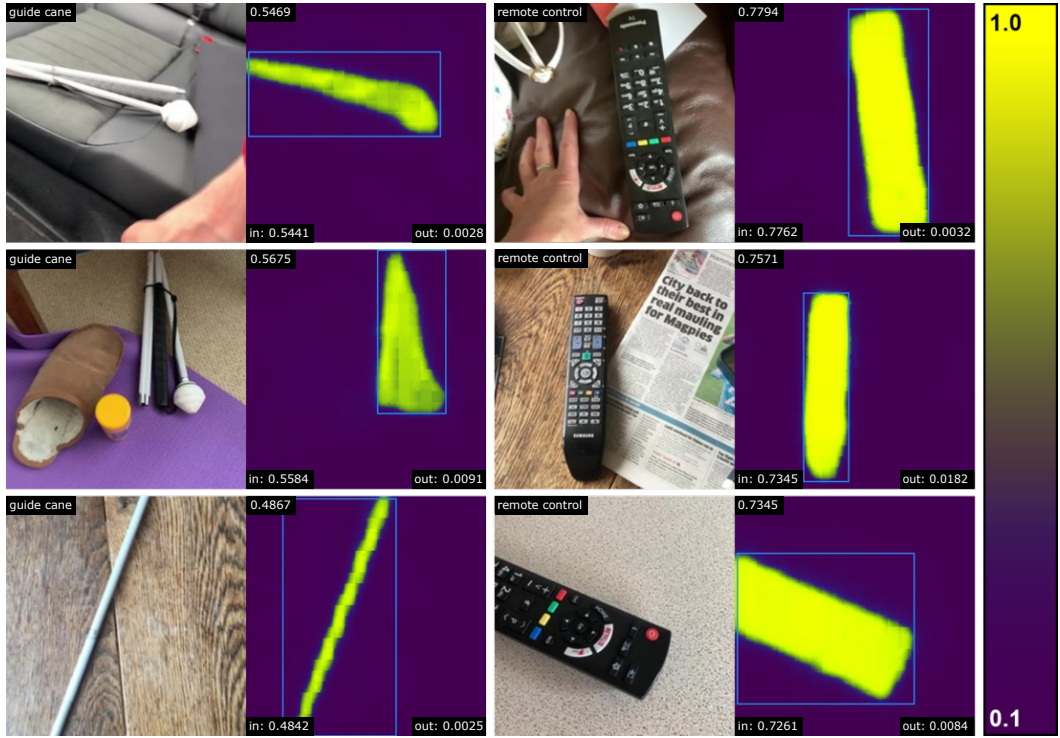}
    \caption{\textbf{CLIPSeg segments non-disability objects (right: TV remote) with higher confidence than disability objects (left: guide cane).} For each image, we report the average confidence score over all pixels inside and outside the object's ground-truth bounding box (``in'' and ``out'', respectively), considering only pixels above a 0.1 confidence threshold. See quantitative results in~\cref{app:tab:image-analysis:seg1}.}
    \label{app:fig:example-analysis:seg1}
\end{figure*}

Semantic segmentation models are also highly likely to be integrated into assistive applications to help \gls{blv} users localize objects.
We examine CLIPSeg~\cite{lueddecke22cvpr} which trains a decoder on top of CLIP's frozen vision and text encoders to enable zero-shot image segmentation from text prompts.
Unlike OWL-ViT, CLIPSeg does not fine-tune the CLIP encoders, and its pre-trained embeddings are used directly.
As before, we run all 180 images through the model with the cleaned noun phrases as text prompts.
We find:

\textbf{Segmentation maps of disability objects are less confident than those of non-disability objects.}
In~\cref{app:tab:image-analysis:seg1}, we compute the average confidence value over all pixels in the segmentation map that fall within the ground-truth bounding box of the target object.
To control for the degree of background present across bounding boxes (especially for irregular-shaped objects), we only consider pixels that have a confidence score greater than 0.1.
With this, we find that CLIPSeg's segmentation maps are $\sim$2x more confident for non-disability objects compared to disability objects.
\begin{table}
    \centering
    \caption{\textbf{CLIPSeg segments non-disability objects with higher confidence than non-disability objects}. The average confidence (with 95\% c.i.) is reported over all pixels with a confidence above 0.1 within the object's ground-truth bounding box.}
    \vspace{-0.5em}
    \scalebox{0.9}{
    \begin{tabular}{l|c}
        \toprule
        & \textbf{Avg in-box confidence} \\
        \midrule
         Disability objects &  0.2181 (0.0842) \\
         Non-disability objects & \textbf{0.4276 (0.1286)}\\
         \bottomrule
    \end{tabular}}
    \label{app:tab:image-analysis:seg1}
    \vspace{-0.5em}
\end{table}
In~\cref{app:fig:example-analysis:seg1}, we show CLIPSeg's segmentations for a guide cane versus a TV remote, two objects for which the confidence score difference was most pronounced.

\textbf{Disability objects are more likely to be segmented as the incorrect object in realistic settings compared to non-disability objects.}
In~\cref{app:tab:image-analysis:seg2}, we compute a confusion score per image: the average confidence score of all the pixels that fall \emph{outside} the object's ground-truth bounding box, divided by the average confidence score over all pixels in the image.
This gives us a measure of how confidently the model is segmenting objects besides the ground-truth object, where a high score indicates the segmentation may be a false positive.
Here we also only include confidence scores above a 0.1 threshold.
We see that CLIPSeg is $\sim$2x more likely to confuse a disability object with another object compared to a non-disability object in ORBIT Clutter images where multiple objects are present.
\begin{table}
    \centering
    \caption{\textbf{CLIPSeg incorrectly segments disability objects more often than non-disability objects on the ORBIT Clutter dataset}. Numbers are the average confidence over all pixels \emph{outside} the object's ground-truth bounding box divided by the average confidence over all pixels in the image (with 95\% c.i.), considering only pixels above a 0.1 confidence threshold.}
    \vspace{-0.5em}
    \scalebox{0.9}{
    \begin{tabular}{l|c}
        \toprule
        & \textbf{Avg confusion score} \\
        \midrule
         Disability objects & \textbf{0.2698 (0.1990)} \\
         Non-disability objects & 0.1292 (0.0749) \\
         \bottomrule
    \end{tabular}}
    \vspace{-1em}
    \label{app:tab:image-analysis:seg2}
\end{table}

\begin{figure*}
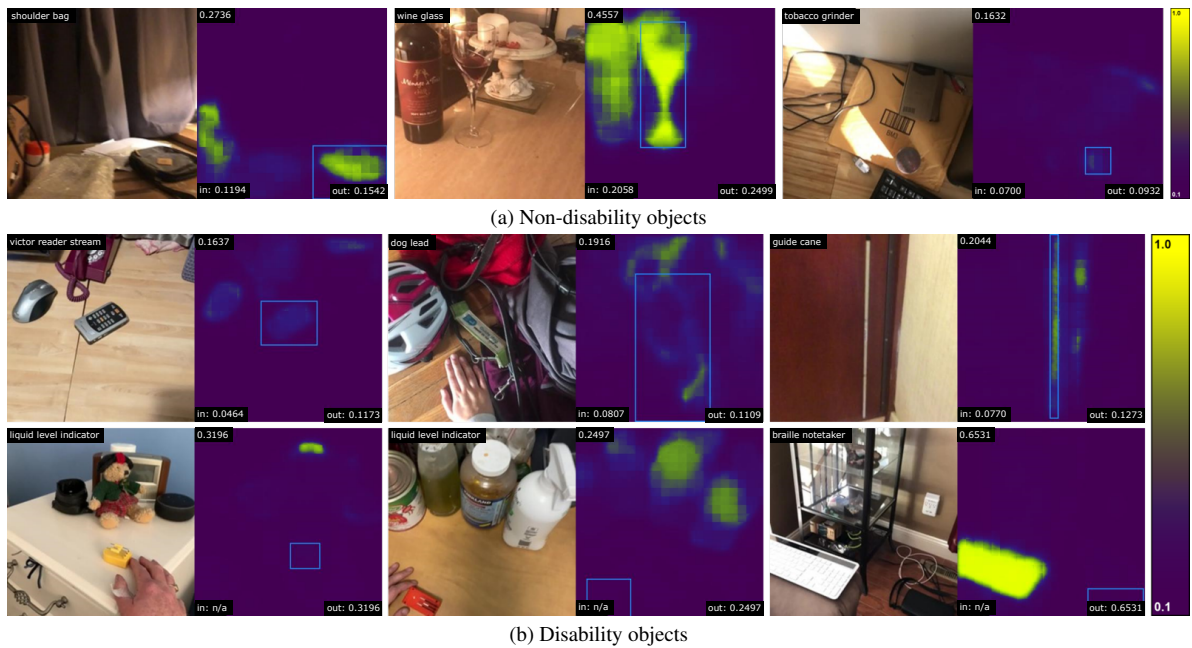

    \centering
    \begin{subfigure}{0.9\textwidth}
        \centering
        \includesvg[width=\textwidth,inkscapelatex=false]{plots/image-analysis-seg2b.svg}
        \caption{Non-disability objects}
        \label{app:fig:example-analysis:seg2a}
    \end{subfigure}
    \begin{subfigure}{0.9\textwidth}
        \centering
        \includesvg[width=\textwidth,inkscapelatex=false]{plots/image-analysis-seg2a.svg}
        \caption{Disability objects}
        \label{app:fig:example-analysis:seg2b}
    \end{subfigure}
    \caption{\textbf{CLIPSeg is more likely to segment disability objects (bottom) incorrectly in cluttered scenes compared to disability objects (top).} For each image, we report the average confidence score over all pixels inside and outside the object's ground-truth bounding box (``in'' and ``out'', respectively), considering only pixels above a 0.1 confidence threshold. The correct object is marked by the bounding box. See quantitative results in~\cref{app:tab:image-analysis:seg2}.}
    \label{app:fig:example-analysis:seg2}
\end{figure*}

We include examples of this in~\cref{app:fig:example-analysis:seg2}.
Here we see that CLIPSeg fails to segment prominent disability objects (see liquid level indicators, guide canes, and Braille notetakers in~\cref{app:fig:example-analysis:seg2b}) but succeeds in segmenting non-disability objects in similarly cluttered scenes (see shoulder bags and wine glasses in~\cref{app:fig:example-analysis:seg2a}).

\subsection{Text-to-image generations with DALL-E2}

\paragraph{Prompt templates.} Three annotators manually created two prompts for each of the 20 objects.
The first prompt was just the cleaned noun phrase (taken from~\cref{app:tab:image-analysis:objects}). 
The second prompt combined the cleaned noun phrase with a surface and up to two adjacent objects.
The surface and adjacent objects were selected to match an image from the ORBIT Clutter dataset of that object.
The image was selected such that it had at least one adjacent object present.
%
%If no images contained adjacent objects, then the first frame from a clutter video was selected.
%
The prompt was then created with the template: ``\emph{\textless object\_name \textgreater{} on \textless surface\textgreater{} next to \textless adjacent-object-1\textgreater{} and \textless adjacent-object-2\textgreater{}}" (\eg ``wine glass on a wooden table next to a bottle of wine and a candle").

\begin{figure*}
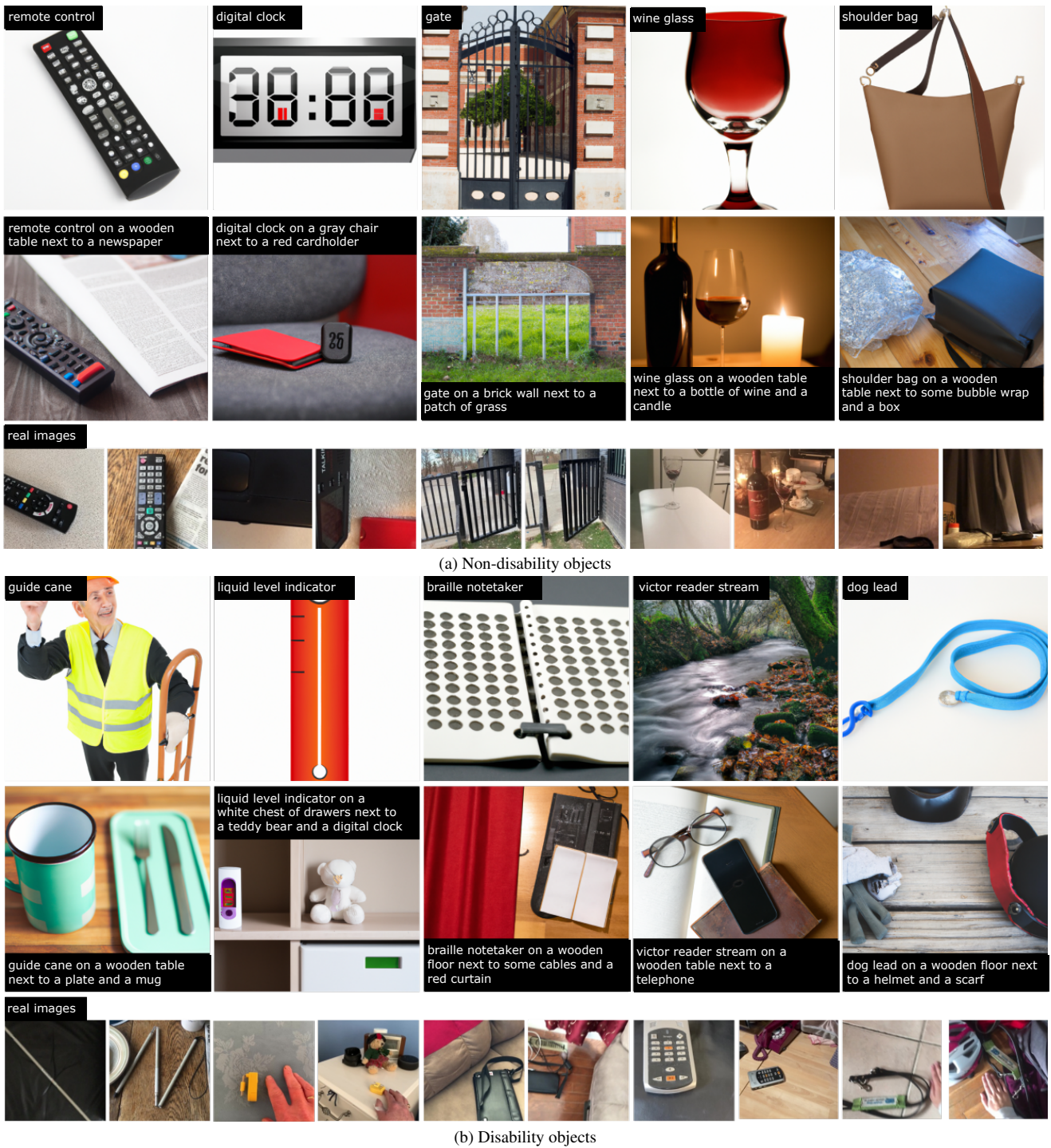

    \centering
    \begin{subfigure}{\textwidth}
        \centering
        \includesvg[width=\textwidth,inkscapelatex=false]{plots/image-analysis-gen1a.svg}
        \caption{Non-disability objects}
        \label{app:fig:image-analysis:gen1a}
    \end{subfigure}
    \begin{subfigure}{\textwidth}
        \centering
        \includesvg[width=\textwidth,inkscapelatex=false]{plots/image-analysis-gen1b.svg}
        \caption{Disability objects}
        \label{app:fig:image-analysis:gen1b}
    \end{subfigure}
    \caption{\textbf{DALL-E2 generates high-quality images of non-disability objects (a), but defaults to more common objects or fabrications for disability objects (b).} For each sub-figure, the top row shows generations for a simple prompt containing just the object name, while the second row shows generations for the richer prompt where a surface and adjacent objects are also specified. The bottom row shows real images of each object.}
    \label{app:fig:image-analysis:gen1}
\end{figure*}

We extend~\cref{fig:image-analysis:gen1} in the main paper with~\cref{app:fig:image-analysis:gen1} here.
We show DALL-E2's generations for the two prompt types for non-disability (\cref{app:fig:image-analysis:gen1a}) and disability (\cref{app:fig:image-analysis:gen1b}) objects.
Overall, we see that DALL-E2 does not generate correct representations for many of the disability objects, either defaulting to a common object or fabricating an object entirely.
In contrast, the generations for non-disability objects are highly realistic and mostly correct.

\end{document}